\documentclass[3p]{elsarticle}
\makeatletter
\def\ps@pprintTitle{%
	\let\@oddhead\@empty
	\let\@evenhead\@empty
	\def\@oddfoot{}%
	\let\@evenfoot\@oddfoot}
\makeatother
\usepackage[bookmarks=false]{hyperref}
    \hypersetup{colorlinks,
      linkcolor=blue,
      citecolor=blue,
      urlcolor=blue}
\usepackage{amsmath}

\usepackage{amssymb}
\usepackage{mathtools}
\usepackage{amsmath}
\usepackage{tikz}
\usepackage{tikz}
\usetikzlibrary{bayesnetLJ}
\usepackage{float}
\usepackage{microtype} 
\usepackage{subfig}
\usepackage[misc]{ifsym}
\usepackage{eurosym}
\usepackage{amsthm}

\usepackage[figuresright]{rotating}
\DeclareMathOperator*{\argmax}{argmax}

\def\bX{{\bf X}}

\def\bS{{\bf S}}

\def\bg{{\bf g}}
\def\ba{{\bf a}}
\def\bd{{\bf d}}

\newcommand\eg{{\em e.g., }}
\newcommand\ie{{\em i.e., }}
\newtheorem{proposition}{Proposition}
\newtheorem{definition}{Definition}
\newcommand{\refextendedversion}{}

\def\btheta{{\boldsymbol\theta}}

\def\bpi{{\boldsymbol\pi}}

\def\balpha{{\boldsymbol\alpha}}
\def\bbeta{{\boldsymbol\beta}}
\def\bgamma{{\boldsymbol\gamma}}
\def\bdelta{{\boldsymbol\delta}}

\begin{document}
\begin{frontmatter}

%% Title, authors and addresses

%% use the tnoteref command within \title for footnotes;
%% use the tnotetext command for the associated footnote;
%% use the fnref command within \author or \address for footnotes;
%% use the fntext command for the associated footnote;
%% use the corref command within \author for corresponding author footnotes;
%% use the cortext command for the associated footnote;
%% use the ead command for the email address,
%% and the form \ead[url] for the home page:
%%
%% \title{Title\tnoteref{label1}}
%% \tnotetext[label1]{}
%% \author{Name\corref{cor1}\fnref{label2}}
%% \ead{email address}
%% \ead[url]{home page}
%% \fntext[label2]{}
%% \cortext[cor1]{}
%% \address{Address\fnref{label3}}
%% \fntext[label3]{}

%\dochead{24th International Conference on Knowledge-Based and Intelligent Information \& Engineering Systems}%
%% Use \dochead if there is an article header, e.g. \dochead{Short communication}
%% \dochead can also be used to include a conference title, if directed by the editors
%% e.g. \dochead{17th International Conference on Dynamical Processes in Excited States of Solids}

\title{Discriminative Viewer Identification using Generative Models of Eye Gaze}

%% use optional labels to link authors explicitly to addresses:
%% \author[label1,label2]{<author name>}
%% \address[label1]{<address>}
%% \address[label2]{<address>}

\author[a]{Silvia~Makowski\corref{cor1}} 
\author[a]{Lena~A.~J\"ager}
\author[b]{Lisa~Schwetlick}
\author[b]{\\Hans~Trukenbrod}
\author[b]{Ralf~Engbert}
\author[a]{Tobias~Scheffer}
%\author[a,b]{Third Author\corref{cor1}}

\address[a]{University of Potsdam, Department of Computer Science, 14482 Potsdam, Germany}
\address[b]{University of Potsdam, Department of Psychology, 14476 Potsdam, Germany}

\begin{abstract}
%% Text of abstract
We study the problem of identifying viewers of arbitrary images based on their eye gaze. 
Psychological research has derived generative stochastic models of eye movements. In order to exploit this background knowledge within a discriminatively trained classification model, we derive Fisher kernels from different generative models of eye gaze. Experimentally, we find that the performance of the classifier strongly depends on the underlying generative model. Using an SVM with Fisher kernel improves the classification performance over the underlying generative model.
\end{abstract}

%\begin{keyword}
%eye movements; eye tracking; Fisher kernel;  

%% keywords here, in the form: keyword \sep keyword

%% PACS codes here, in the form: \PACS code \sep code

%% MSC codes here, in the form: \MSC code \sep code
%% or \MSC[2008] code \sep code (2000 is the default)

%\end{keyword}
\cortext[cor1]{Corresponding author. E-mail address: silvia.makowski@uni-potsdam.de}
\end{frontmatter}

%\correspondingauthor[*]{Corresponding author. Tel.: +0-000-000-0000 ; fax: +0-000-000-0000.}
%\email{author@institute.xxx}

%%
%% Start line numbering here if you want
%%
% \linenumbers

%% main text

%\enlargethispage{-7mm}

\section{Introduction}
\label{sec:introduction}
Human eye movements are driven by a highly-complex interplay between voluntary and involuntary processes related to oculomotor control, high-level vision, cognition, and attention. While exploring a scene, the eyes move their focus three to four times per second on average by performing very fast movements, termed saccades \cite{henderson1998eye}. This type of active perception is functional, since high visual acuity is only obtained within the fovea, a very small area on the retina. Visual uptake is limited to phases of relative gaze stability between the saccades, denoted as fixations \citep{henderson1998eye}. The sequence of saccades and fixations that constitute the eye's response to a scene is referred to as {\em scanpath}. 
It has long been known that the way we move our eyes in response to a given stimulus is highly individual~\cite{Noton1971} and more recent psychological research has shown that these individual characteristics are reliable over time~\cite{Bargary2017}. Hence, it has been proposed to use eye movements as a behavioral biometric characteristic~\cite{KasprowskiOber2004,Bednarik2005}. 

%Psychological research has shown that eye movement patterns exhibit a high degree of individual variability in various kinds of visual tasks. 
%such as scene viewing \citep{Andrews1999,Rayner2007,Castelhano2008,Henderson2014}, face processing \citep{Rayner2007,Castelhano2008,Peterson2013}, visual search \citep{Andrews1999,Rayner2007,Boot2009}, or reading \citep{erdmann1898,huey1908,rayner1998eye,Rayner2007,Henderson2014,Afflerbach2015}. 
%It has further been shown that these individual characteristics often persist across different tasks \citep{Andrews1999,Rayner2007,Castelhano2008,Henderson2014} and are stable  over time \citep{Henderson2014,Bargary2017}. 
%These reliable individual differences makes eye movements interesting for biometrics, where they might offer an unobtrusive identification procedure. 
%
Psychologists have developed generative stochastic models in order to explain various aspects of scanpaths. The {\sl SceneWalk} model~\citep{engbert2015spatial} generates saccade amplitudes and directions of a viewer watching an image. A probabilistic model of reading~\citep{Landwehr2014} generates fixation durations, saccade amplitudes and durations, and the types of saccades (regressions to a previous word, refixations of the current word, skips ahead) that can occur during reading.
Generative models constitute background knowledge about eye gaze, but they are optimized to maximize the likelihood of the observed scanpaths, rather than the accuracy of a discriminative task such as viewer identification. 
Fisher kernels allow the use of generative stochastic models as background knowledge to derive a feature representation from sequential data.  
For reading, an SVM with a Fisher kernel derived from a generative model of eye movements has been observed to performs substantially better at reader identification than the generative stochastic model itself~\cite{Makowski2018}. This finding motivates our study on general scene viewing: starting from the {\sl SceneWalk} model~\citep{engbert2015spatial} and from a generative model for reading~\citep{Landwehr2014} which we adapt to general scene viewing and which we extend by incorporating additional features, we derive Fisher kernels that encode scanpaths in terms of their gradient for the generative stochastic models. 

This paper is organized as follows. Section \ref{sec:ProblemSetting} defines the problem setting of viewer identification. Section \ref{sec:GenerativeModels} introduces two generative models of scanpaths, an adaptation of a reader identification model \citep{Landwehr2014} and the SceneWalk model \citep{engbert2015spatial}. In Section \ref{Sec:Fisher}, we develop the Fisher kernel function from these models. In Section \ref{sec:EmpiricalStudy}, we evaluate our model and several baseline models. Section \ref{sec:Conclusion} concludes.

\section{Problem Setting}
\label{sec:ProblemSetting}
When exploring a scene presented on a screen, a viewer generates a scanpath, which is a sequence
\mbox{$\bS = ((q_1, d_1),\ldots,(q_T,d_T))$} of fixation positions $q_t$, measured in degrees of visual angle, and fixation durations $d_t$, measured in milliseconds. %Scanpaths can be observed precisely with an eye-tracking system.
We study the problem of viewer identification and therefore train a model that selects the conjectured identity $y$ of a viewer that generates a scanpath $\bS$ on a certain picture, from a set of individuals that are known at training time.
Training data consists of a set $\mathcal{D} = \{(\bS_1,\bX_1,y_1),...,(\bS_n,\bX_n,y_n)\}$
of scanpaths $\bS_1,...,\bS_n$ that have been obtained from subjects viewing pictures $\bX_1,...,\bX_n$, labeled with viewers' identities $y_1,...,y_n$. 

\section{Generative Models of Scanpaths}
\label{sec:GenerativeModels}
Let $p(\mathbf{S}|\mathbf{X},\btheta)$ be a parametric model of scanpaths given a picture $\bX$. 
%Given a set of scanpaths and images ${\mathcal{\bar D}} = \{(\bS_i,\bX_i)\}$, 
%model parameters can be estimated by maximum likelihood. 
In a generative setting, viewer-specific models $p(\mathbf{S}|\mathbf{X},\btheta_y)$ for user $y$ can be estimated on viewer-specific data ${\mathcal{\bar D}}_y = \{(\bS_i,\bX_i)|(\bS_i,\bX_i,y_i)\in {\mathcal{D}}, y_i=y\}$ by maximum likelihood.
At application time, the prediction for a scanpath $\bS$ on a new picture $\bX$ can be obtained as $y^* = \argmax_y p(\mathbf{S}|\mathbf{X},\btheta_y)$.
For the discriminative setting we develop in Section~\ref{Sec:Fisher}, generative parameters are estimated on all training data ${\mathcal{\bar D}} = \{(\bS_i,\bX_i)|(\bS_i,\bX_i,y_i)\in {\mathcal{D}}\}$, and a Fisher score representation is derived from this generative model. 

In this section, we modify a model for reader identification~\cite{Landwehr2014} to the case of viewer identification, and add velocity- and acceleration-based features to this model. We then review the {\sl SceneWalk} model~\citep{engbert2015spatial}.
In Section~\ref{Sec:Fisher}, we will derive the Fisher kernel for both models and thus build a discriminative classifier for viewer identification.

\subsection{Markov Model for Scene Viewing}
\label{sec:AdaptedLandwehr}
%\cite{Landwehr2014} develop a model for reader identification---a problem that differs from viewer identification only in the type of stimulus. 
In this section, we will review and adapt a model for reader identification~\cite{Landwehr2014} to reflect how viewers generate fixations while exploring a picture. 
The model assumes that the joint distribution over all fixation positions and durations is created by a Markov process:
\begin{equation}
\label{eq: Landwehr Joint Distribution}
p(q_1,\dots,q_T,d_1,\dots,d_T|\bX,\btheta) 
=p(q_1,d_1|\bX,\btheta)\prod_{t=1}^{T-1} p(q_{t+1},d_{t+1}|q_{t},\bX,\btheta);
\end{equation}
for this reason, we will refer to the model as {\sl Markov model} in the following.
To model the conditional distribution $p(q_{t},d_{t}|q_{t-1},\bX,\btheta)$ of the next fixation position $q_{t}$ and duration $d_{t}$ given the current fixation position $q_{t}$, the original model distinguishes between the {\em saccade types} of {\em regression} to a previous word, {\em refixation} of the current word before or after the current position, fixation of the {\em next word} or {\em skipping} one or more words. Our adaptation of the model distinguishes four saccade types $u$: the scanpath can {\em maintain} the direction of the previous saccade up to $\pm 45^\circ$ ($u=1$), change saccade direction to the {\em right} ($u=2$), or to the {\em left} ($u=3$) by more than $45^\circ$, or {\em reverse} direction by turning between $135^\circ$ and $225^\circ$ ($u=4$). 
At time $t$, the model first draws a saccade type
%\begin{equation}
%\label{eq: Landwehr Saccade Type}
$u_{t} \sim p(u|\bpi)= \mathrm{Mult}(u|\bpi)$ %\\
%\end{equation}
from a multinomial distribution. Both the original and adapted models then draw a saccade amplitude $a_{t} \sim p(a| u_{t})$,  measured as the change of degrees of visual angle, from type-specific gamma distributions
%\begin{align}&
$p(a|u_{t}=u,\balpha^a,\bbeta^a) = \mathcal{G}(a|\alpha_u^a,\beta_u^a)$ for $u \in \{1,...,4\}$, %\label{eq:amplitude_positive}
%\end{align}
where $\balpha^a=\{\alpha_u^a|u \in \{1,...,4\}\}$, $\bbeta^a=\{\beta_u^a|u \in \{1,...,4\}\}$ and $\mathcal{G}(\cdot|\alpha^a,\beta^a)$ is the gamma distribution parameterized by
shape $\alpha^a$ and scale $\beta^a$.
Analogously, the model draws a fixation duration $d_{t} \sim p(d| u_{t},\balpha^d,\bbeta^d)$, also from type-specific gamma distributions
%\begin{align}
$p(d|u_{t}=u,\balpha^d,\bbeta^d) = \mathcal{G}(d|\alpha_u^d,\beta_u^d)$ for $u \in \{1,...,4\}$, %\label{eq:duration}
%\end{align}
where $\balpha^d=\{\alpha_u^d|u \in \{1,...,4\}\}$ and $\bbeta^d=\{\beta_u^d|u \in \{1,...,4\}\}$.

\subsubsection{Parameter Estimation}
Given a set of $k$ scanpaths on images ${\mathcal{\bar D}} = \{(\bS_i,\bX_i)\}$, all parameters are aggregated into a vector $\btheta$ and estimated by optimizing a maximum likelihood criterion
%\begin{equation}
%\label{eq:maximum_likelihood_landwehr}
$\btheta^* = \argmax_{\btheta} \sum_{i=1}^k \ln p(\bar\bS_i|\bar\bX_i,\btheta)$.
%\end{equation}
Given $\bar{\mathcal{D}}$, all fixation positions $q_t$ and saccade types $u_t$ are known and the likelihood factorizes into separate likelihood terms depending on saccade type, amplitude, and duration parameters:
\begin{multline}
\btheta^* = \argmax_{\bpi,\balpha^a,\bbeta^a,\balpha^d,\bbeta^d} \bigg(\sum_{i=1}^k \sum_{t=1}^{T_i} \ln \mathrm{Mult}(u^{(i)}_t| \bpi) 
+ \sum_{i=1}^k \sum_{t=1}^{T_i} \ln p(a^{(i)}_t | u^{(i)}_t,\balpha^a,\bbeta^a) 
+ \sum_{i=1}^k \sum_{t=1}^{T_i} \ln p(d^{(i)}_t | u^{(i)}_t,\balpha^d,\bbeta^d) \bigg). 
\label{eq:likelihood_factorized_landwehr}
\end{multline}

\subsection{Markov Model with Saccade Dynamics}\label{sec:AdaptedLandwehr2}
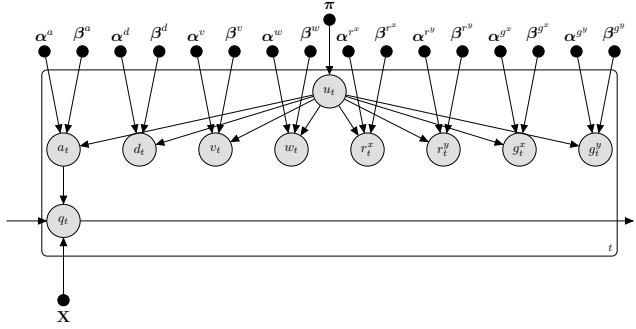
\begin{figure}
	\centering
	\scalebox{0.5}{
		\begin{tikzpicture}
		\node[obs, minimum size=25pt] (u1) {$u_t$};
		\node[obs, below=of u1, minimum size=25pt, xshift=-7cm,  yshift=10] (a1) {$a_t$};
		\node[obs, below=of u1, minimum size=25pt, xshift=-5cm, yshift=10] (d1) {$d_t$};
		\node[obs, below=of u1, minimum size=25pt, xshift=-3cm,  yshift=10] (v1) {$v_t$};
		\node[obs, below=of u1, minimum size=25pt, xshift=-1cm, yshift=10] (w1) {$w_t$};
		\node[obs, below=of u1, minimum size=25pt, xshift=1cm, yshift=10] (rx1) {$r^x_t$};
		\node[obs, below=of u1, minimum size=25pt, xshift=3cm, yshift=10] (ry1) {$r^y_t$};
		\node[obs, below=of u1, minimum size=25pt, xshift=5cm, yshift=10] (gx1) {$g^x_t$};
		\node[obs, below=of u1, minimum size=25pt, xshift=7cm, yshift=10] (gy1) {$g^y_t$};
		\node[obs,below=of a1, minimum size=25pt] (q1) {$q_t$};
		\node [const, above=of a1,  yshift=1cm, xshift=-0.5cm, label=above:\large$\boldsymbol\alpha^a$] (alphaa) {};
		\node [const, above=of a1,  yshift=1cm, xshift=0.5cm, label=above:\large$\boldsymbol\beta^a$] (betaa) {};
		\node [const, above=of d1,  yshift=1cm, xshift=-0.5cm, label=above:\large$\boldsymbol\alpha^d$] (alphad) {};
		\node [const, above=of d1,  yshift=1cm, xshift=0.52cm, label=above:\large$\boldsymbol\beta^d$] (betad) {};
		\node [const, above=of v1,  yshift=1cm, xshift=-0.5cm, label=above:\large$\boldsymbol\alpha^v$] (alphav) {};
		\node [const, above=of v1,  yshift=1cm, xshift=0.5cm, label=above:\large$\boldsymbol\beta^v$] (betav) {}; 
		\node [const, above=of w1,  yshift=1cm, xshift=-0.5cm, label=above:\large$\boldsymbol\alpha^w$] (alphaw) {};
		\node [const, above=of w1,  yshift=1cm, xshift=0.5cm, label=above:\large$\boldsymbol\beta^w$]  (betaw) {};
		\node [const, above=of rx1,  yshift=1cm, xshift=-0.5cm, label=above:\large$\boldsymbol\alpha^{r^x}$] (alpharx) {};
		\node [const, above=of rx1,  yshift=1cm, xshift=0.5cm, label=above:\large$\boldsymbol\beta^{r^x}$]  (betarx) {};
		\node [const, above=of ry1,  yshift=1cm, xshift=-0.5cm, label=above:\large$\boldsymbol\alpha^{r^y}$] (alphary) {};
		\node [const, above=of ry1,  yshift=1cm, xshift=0.5cm, label=above:\large$\boldsymbol\beta^{r^y}$]  (betary) {};
		\node [const, above=of gx1,  yshift=1cm, xshift=-0.5cm, label=above:\large$\boldsymbol\alpha^{g^x}$] (alphagx) {};
		\node [const, above=of gx1,  yshift=1cm, xshift=0.5cm, label=above:\large$\boldsymbol\beta^{g^x}$]  (betagx) {};
		\node [const, above=of gy1,  yshift=1cm, xshift=-0.5cm, label=above:\large$\boldsymbol\alpha^{g^y}$] (alphagy) {};
		\node [const, above=of gy1,  yshift=1cm, xshift=0.5cm, label=above:\large$\boldsymbol\beta^{g^y}$]  (betagy) {};
		\node [const, above=of u1,  yshift=0.3cm, label=above:\large$\boldsymbol\pi$] (pi) {};
		\node[const, below=of q1,  yshift=-0.5cm,  label=below:\large$\mathbf X$] (X) {};
		\plate  {plate1}  {(u1) (q1) (a1) (d1) (v1) (w1) (rx1) (ry1) (gx1) (gy1)} {$t$}; 
		\edge {u1} {d1,a1,v1,w1,rx1,ry1,gx1,gy1} 
		\edge {a1} {q1} 
		\edge{alphaa}{a1}
		\edge{betaa}{a1}
		\edge{alphad}{d1}
		\edge{betad}{d1}
		\edge{alphav}{v1}
		\edge{betav}{v1}
		\edge{alphaw}{w1}
		\edge{betaw}{w1}
		\edge{alpharx}{rx1}
		\edge{betarx}{rx1}
		\edge{alphary}{ry1}
		\edge{betary}{ry1}
		\edge{alphagx}{gx1}
		\edge{betagx}{gx1}
		\edge{alphagy}{gy1}
		\edge{betagy}{gy1}    
		\edge{pi}{u1}
		\edge{X}{q1}
		\draw[<-] (q1) --++(-1.5cm,0);
		\draw[->] (q1) --++(0:15cm);
		\end{tikzpicture}
	}
	\caption{Plate notation of the {\sl Markov model with saccade dynamics}.}
	\label{fig:graphical_model}
\end{figure}

Prior work on biometric identification using eye gaze has shown that  saccade velocities, acceleration~\citep{Holland2013}, and the relationship between peak velocity and amplitude of a saccade---referred to as {\em vigor}---convey information about viewer identity~\citep{Rigas2016}. We therefore further extend the {\sl Markov model} to include features that describe the saccade dynamics; we will study whether modeling these attributes of scanpaths contributes to identification accuracy. 
A survey of the data set used for evaluation in Section~\ref{sec:EmpiricalStudy} shows that mean saccade velocities and accelerations follow Gamma distributions. Therefore, we extend the model to draw saccade mean velocities $v$ as in Equation~\ref{eq:velocity_positive} and mean accelerations as in Equation~\ref{eq:mean_acceleration_positive} from type-specific gamma distributions:
\begin{align}
v_{t}&\sim p(v|u_{t}=u,\balpha^v,\bbeta^v) = \mathcal{G}(v|\alpha_u^v,\beta_u^v) \text{ for $u \in \{1,...,4\}$} \label{eq:velocity_positive}\\
w_{t}&\sim p(w|u_{t}=u,\balpha^w,\bbeta^w) = \mathcal{G}(w|\alpha_u^w,\beta_u^w) \text{ for $u \in \{1,...,4\}$}. \label{eq:mean_acceleration_positive}
\end{align}
We define the peak acceleration-to-deceleration ratio of the horizontal saccade vector component $r^x_{t}$ of saccade $t$ as the ratio of the horizontal peak acceleration divided by the horizontal peak deceleration; the vertical peak acceleration-to-deceleration ratio $r^y_{t}$ is defined in analogy; both ratios are governed by Gamma distributions: 
\begin{align}
r^{x}_{t}&\sim p(r^x|u_{t}=u,\balpha^{r^x},\bbeta^{r^x}) = \mathcal{G}(r^x|\alpha_u^{r^x},\beta_u^{r^x}) \text{ for $u \in \{1,...,4\}$}\\ \label{eq:acceleration_ratiox}
r^{y}_{t}&\sim p(r^y|u_{t}=u,\balpha^{r^y},\bbeta^{r^y}) = \mathcal{G}(r^y|\alpha_u^{r^y},\beta_u^{r^y}) \text{ for $u \in \{1,...,4\}$}. 
\end{align}
%We model the relationship between peak velocity $v^{max}_{t}$ and amplitude $a_{t}$ of a saccade $t$, referred to as {\em vigor} $g_{t}$.
The relationship between the peak velocity $v^{max}_{t}$, amplitude $a_{t}$, and vigor $g_{t}$ of a saccade $t$ follows a parametric relationship $v^{max}_{t} = g_{t}\left(1-e^{\frac{-a_{t}}{b}}\right)$~\cite{baloh1975quantitative} with a global scalar rate parameter $b$.
%\begin{align}
%v^{max}_{t} = g_{t}\left(1-e^{\frac{-a_{t}}{b}}\right).
%\label{eq:peak_vel_amplitude}
%\end{align} 
Following Rigas et al.~\cite{Rigas2016}, we fit rate parameter $b$ in two steps. First, $b$ and all $g_{t}$ are jointly estimated via least-squares fitting 
%of Equation~\ref{eq:peak_vel_amplitude} 
on saccadic training data for each subject separately; then values $b$ are averaged across subjects into a global rate parameter $b^*$. 
%Then the vigor $g$ for each saccade is calculated by solving Equation~\ref{eq:peak_vel_amplitude} for $g$:
%$$g = v^{\text{peak}}/(1-\exp(-a/b_{\text{avg}}))$$ 
We incorporate the saccadic vigor for the vertical and horizontal components of each saccade and into the generative model via gamma distributions:
\begin{align}
g^{x}_{t}&\sim p(g^x|u_{t}=u,\balpha^{g^x},\bbeta^{g^x}) = \mathcal{G}(g^x|\alpha_u^{g^x},\beta_u^{g^x}) \text{ for $u \in \{1,...,4\}$}\\ 
g^{y}_{t}&\sim p(g^y|u_{t}=u,\balpha^{g^y},\bbeta^{g^y}) = \mathcal{G}(g^y|\alpha_u^{g^y},\beta_u^{g^y}) \text{ for $u \in \{1,...,4\}$}. 
\end{align}
Figure~\ref{fig:graphical_model} shows the {\sl Markov model with saccade dynamics} as a plate diagram.

\subsubsection{Parameter Estimation}
\label{sec:parameter_estimation_saccade_dynamics}
Apart from the global rate parameter $b$, all model parameters are fitted for each user separately via maximum likelihood. The likelihood function is detailed in Appendix~\ref{ap:xlandwehr}\refextendedversion.

\subsection{The SceneWalk Model}\label{sec:SceneWalk}
{\sl SceneWalk}~\citep{engbert2015spatial} assumes the joint distribution over all fixation positions and durations of a scanpath to factorize as:  
\begin{align}
\label{eq: SceneWalk Distribution New}
p(q_1,\dots,q_T|\bX,\btheta) = p(q_1|\bX,\btheta)\prod_{t=1}^{T-1} p(q_{t+1}|q_{1},...,q_{t},d_{1},...,d_{t},\bX,\btheta).
\end{align}
Here, $p(q_1|\bX,\btheta)$ is the likelihood of the first fixation position and can either be given by the experimental design (\eg by a fixation cross at a certain position that triggers the onset of an image) or the model itself~\citep{schutt2017likelihood}.

When position $q_t= (i_t, j_t)$ is fixated, % with duration $d_t$, 
the model assigns a potential to each image pixel $(i,j)$ to be the next saccade target $q_{t+1}$. This potential is obtained from an attentional component $\mathbf{A}_{t}$ and from an inhibitory component $\mathbf{F}_{t}$.
Both components are based on Gaussian windows $\mathbf{G}_{t}^{A}$ and $\mathbf{G}_{t}^{F}$, respectively, centered at the position of $q_t$ and with standard deviations $\sigma_A$ and $\sigma_F$:
\begin{equation}
\mathbf{G}_{t}^{A/F}(i,j) = \frac{1}{2 \pi \sigma_{A/F}^2} \exp\left(\frac{(i-i_t)^2 + (j-j_t)^2}{2 \sigma_{A/F}^2}\right).
\end{equation}

\subsubsection{Attentional Component}
The attentional component refers to the empirical {\em saliency} $\mathbf{H}$ of the image. The saliency map characterizes the intrinsic potential $\mathbf{H}(i,j)$ of image positions $(i,j)$ to attract visual attention. The saliency is time-independent and can be obtained for each image separately, but globally across all viewers. It is common practice to estimate the saliency by kernel density estimation with a bandwidth determined by Scott's Rule~\citep{baddeley2015,engbert2015spatial}. The  attentional component $\mathbf{A}_{t}$ is a dynamically evolving matrix that accesses the saliency matrix through a Gaussian window $\mathbf{G}_{q_t}^{A}$ which simulates the foveal area of high-acuity vision. The attentional component (Equation~\ref{eq:16}) changes over time at a rate of $\omega_A$. 
\begin{equation}
\mathbf{A}_{t} = \frac{\mathbf{G}_{t}^A \mathbf{H}}{\sum_{i,j} \mathbf{G}_{t}^A(i,j) \mathbf{H}(i,j)} + e^{-\omega_A d_t} \left(\mathbf{A}_{{t-1}} - \frac{\mathbf{G}_{t}^A \mathbf{H}}{\sum_{i,j} \mathbf{G}_{t}^A(i,j) \mathbf{H}(i,j)}\right)\label{eq:16}
\end{equation}
\subsubsection{Inhibitory Component}
The inhibitory component $\mathbf{F}_{t}$ uses its Gaussian window to build up inhibition around the current fixation position and thus provokes an exploration of new regions of the image. It changes over time with a rate $\omega_F$ as in Equation~\ref{eq:17}.
\begin{equation}
\mathbf{F}_{t} = \frac{\mathbf{G}_{t}^F}{\sum_{i,j} \mathbf{G}_{t}^A(i,j)} + e^{-\omega_F d_t} \left(\mathbf{F}_{t-1} - \frac{\mathbf{G}_{t}^F}{\sum_{i,j} \mathbf{G}_{t}^F(i,j) }\right)\label{eq:17}
\end{equation}
Both, the attentional and the inhibitory component, are calculated recursively, since they need the respective components of the previous fixation $q_{t-1}$. 

\subsubsection{Combined Potential for Target Selection}
In the {\sl SceneWalk} model, parameter $c_F$ trades the attentional against the inhibitory component; $\lambda$ and $\gamma$ serve as regularization parameters. Equation~\ref{eq:Ut} shows the resulting potential $\mathbf{U}_{t}$ for target selection.
\begin{equation}
\mathbf{U}_{t} = \frac{\mathbf{A}_{t}^\lambda}{\sum_{i,j}\mathbf{A}_{t}(i,j)^\lambda} - c_F \frac{\mathbf{F}_{t}^\gamma}{\sum_{i,j}\mathbf{F}_{q}(i,j)^\gamma}\label{eq:Ut}
\end{equation}

\subsubsection{Probabilities of image positions}
Given a scanpath of fixation positions $q_{1},...,q_{t}$ and durations $d_{1},...,d_{t}$, the model calculates a probability for each possible image position to be the next fixation position $q_{t+1}$ in the scanpath as a mixture of the normalized potential $\mathbf{U}_{q_t}$ and the uniform distribution over all image positions $(i,j)$, weighted by a regularization parameter $\zeta \in [0,1]$:
\begin{align}
p(q_{t+1}|q_1,...,q_{t}, d_1,...,d_{t}, \theta) 
= (1-\zeta) \frac{\mathbf{U}_{t}(i_{t+1}, j_{t+1})}{\sum_{i,j} \mathbf{U}_{t}(i, j)} + \zeta \frac{1}{\sum_{i,j} 1}.
\end{align}

\subsubsection{Parameter Estimation}
In total, the parameter vector $\btheta$ of {\sl SceneWalk} consists of eight parameters  $\omega_A$, $\omega_F$, $\sigma_A$, $\sigma_F$, $\gamma$, $\lambda$, $c_F$, $\zeta$.
While \cite{engbert2015spatial} fit these parameters using maximum likelihood, we find this estimation technique to be numerically unstable and therefore resort to 
maximizing a regularized maximum likelihood criterion 
%\begin{equation}
$\btheta^* = \argmax_{\btheta} \sum_{i=1}^k \ln p(\bar\bS_i|\bar\bX_i,\btheta) - \rho \sum_j \theta_j^2$.
%\end{equation}

\section{Fisher Kernel}
\label{Sec:Fisher}
Fisher kernels~\citep{jaakkola1999} are a common framework that exploits generative probabilistic models as a representation of sequential or other structured instances within discriminative classifiers. The Fisher kernel approach projects structured input---here, scanpaths---into the gradient space of a generative probability model that was previously fitted to the training data via maximum likelihood. This section derives Fisher representations based on the generative models described in Sections \ref{sec:AdaptedLandwehr}, \ref{sec:AdaptedLandwehr2}, and \ref{sec:SceneWalk} to map scanpaths into feature vectors. 

\subsection{Fisher Kernel Function}
The Fisher kernel function calculates the similarity of two scanpaths $\mathbf{S}_i$ and $\mathbf{S}_j$ as the inner product in the Riemannian manifold given by the class of probability models.
\begin{definition}[Fisher kernel function]
	Let $\btheta^*$ be the maximum likelihood estimate of a generative model on all training data.
	Let $\mathbf{S}_i$, $\mathbf{S}_j$ denote scanpaths on pictures $\bX_i$, $\bX_j$. The Fisher kernel between $\bS_i$, $\bS_j$ is
	%\begin{equation*}
	$K((\bS_i,\bX_i),(\bS_j,\bX_j) = \bg_i^\top \mathbf{I}^{-1} \bg_j$
	%\end{equation*}
	where $\bg_i = \left. \nabla_{\btheta} p(\bS_i|\bX_i,\btheta) \right|_{\theta=\theta^*}$ and where we employ the empirical version of the Fisher information 
	matrix given by $\mathbf{I} = \frac{1}{N} \sum_{i = 1}^N \mathbf{g}_i \mathbf{g}_i^\top.$
\end{definition}
The gradients of the log-likelihood functions of the respective models are derived in Propositions~\ref{prop:gradient_markov}, \ref{prop:gradient_markov_dynamic}, and 3.

\subsection{Fisher Kernel for Markov Model}
\begin{proposition}[Gradient of log-likelihood of the Markov Model] \label{prop:gradient_markov}
	Let $\bS = ((q_1, d_1),\ldots,(q_T,d_T))$ denote a scanpath obtained on an image $\bX$. 
	Let $a_1,...,a_T$ denote the saccade amplitudes, and $u_1,...,u_T$ denote the saccade types in $\bS$. 
	Define for $u \in \{1,2,3,4\}$ the set $\{i^{(u)}_1,...,i^{(u)}_{K_u}\} = \{i \in \{1,...,T\}| u_i = u \}$.
	Let \mbox{$\ba_u = (|a_{i^{(u)}_1}|,...,|a_{i^{(u)}_{K_u}}|)^\top$}, \mbox{$\bd_u = (d_{i^{(u)}_1},...,d_{i^{(u)}_{K_u}})^\top$}.
	Then the gradient of the logarithmic likelihood of the model defined in Section~\ref{sec:AdaptedLandwehr} is 
	{\small\begin{equation*}
	\bg=\nabla_{\btheta}  \ln p(\bS|\bX,\btheta) = (\bar\bg_1^\top,\bar\bg_2^\top,\bar\bg_3^\top,\bar\bg_4^\top)^\top \textrm{, where for } u \in \{1,2,3,4\}:\quad
	%\end{equation*} 
	%{\small\begin{equation*}
		\bar\bg_u =\begin{pmatrix}
		\pi_u^{-1} K_u\\
		\sum\nolimits_{\substack{1 \leq t \leq T:u_t=u}} \ln(a_t) - \psi(\alpha_u^a)- \beta_u^a\\
		\frac{1}{\beta_u^a} \sum\nolimits_{\substack{1 \leq t \leq T:u_t=u}} \Big(\frac{a_t}{\beta_u^a} -\alpha_u^a \Big)\\
		\sum\nolimits_{\substack{1 \leq t \leq T:u_t=u}} \ln(d_t) - \psi(\alpha_u^d)- \beta_u^d\\
		\frac{1}{\beta_u^d} \sum\nolimits_{\substack{1 \leq t \leq T:u_t=u}} \Big(\frac{d_t}{\beta_u^d} -\alpha_u^d \Big)
		\end{pmatrix}.
		\end{equation*}}
\end{proposition}
A proof of Proposition~\ref{prop:gradient_markov} is given in Appendix~\ref{sec:Proof_Markov}\refextendedversion.
\fussy

\subsection{Fisher Kernel for Markov Model with Saccade Dynamics}
\begin{proposition}[Gradient of log-likelihood of the Markov Model with Saccade Dynamics] \label{prop:gradient_markov_dynamic}
	In addition to Proposition~\ref{prop:gradient_markov} for the Markov Model for SceneViewing, let $v_1,...,v_T$ denote the saccade mean velocities and $w_1,...,w_T$ denote the saccade mean accelerations. Let $r_1^x,...,r_T^x$ denote the horizontal and $r_1^y,...,r_T^y$ the vertical peak-acceleration-to-deceleration ratio. Let $g_1^x,...,g_T^x$ denote the horizontal and $g_1^y,...,g_T^y$ the vertical saccade vigor in $\bS$. 
	Then the gradient of the logarithmic likelihood of the model defined in Appendix~\ref{ap:xlandwehr}~\refextendedversion~is 
	\begin{equation*}
	\bg=\nabla_{\btheta}  \ln p(\bS|\bX,\btheta) = (\bar\bg_1^\top,\bar\bg_2^\top,\bar\bg_3^\top,\bar\bg_4^\top)^\top \textrm{, where for } u \in \{1,2,3,4\}:\quad
%	\end{equation*}
	{\scriptsize	%\begin{equation*}
		\bar\bg_u =\begin{pmatrix}
		\pi_u^{-1} K_u\\
		\sum\nolimits_{\substack{1 \leq t \leq T:u_t=u}} \ln(a_t) - \psi\alpha_u^a)- \beta_u^a\\
		\frac{1}{\beta_u^a} \sum\nolimits_{\substack{1 \leq t \leq T:u_t=u}} \Big(\frac{a_t}{\beta_u^a} -\alpha_u^a \Big)\\
		\sum\nolimits_{\substack{1 \leq t \leq T:u_t=u}} \ln(d_t) - \psi(\alpha_u^d)- \beta_u^d\\
		\frac{1}{\beta_u^d} \sum\nolimits_{\substack{1 \leq t \leq T:u_t=u}} \Big(\frac{d_t}{\beta_u^d} -\alpha_u^d \Big)\\
		\sum\nolimits_{\substack{1 \leq t \leq T:u_t=u}} \ln(v_t) - \psi(\alpha_u^v)- \beta_u^v\\
		\frac{1}{\beta_u^v} \sum\nolimits_{\substack{1 \leq t \leq T:u_t=u}} \Big(\frac{v_t}{\beta_u^v} -\alpha_u^v \Big)\\
		\sum\nolimits_{\substack{1 \leq t \leq T:u_t=u}} \ln(w_t) - \psi(\alpha_u^w)- \beta_u^w\\
		\frac{1}{\beta_u^w} \sum\nolimits_{\substack{1 \leq t \leq T:u_t=u}} \Big(\frac{w_t}{\beta_u^w} -\alpha_u^w \Big)\\
		\sum\nolimits_{\substack{1 \leq t \leq T:u_t=u}} \ln(r_t^x) - \psi(\alpha_u^{r^x})- \beta_u^{r^x}\\
		\frac{1}{\beta_u^{r^x}} \sum\nolimits_{\substack{1 \leq t \leq T:u_t=u}} \Big(\frac{r_t^x}{\beta_u^{r^x}} -\alpha_u^{r^x} \Big)\\
		\sum\nolimits_{\substack{1 \leq t \leq T:u_t=u}} \ln(r_t^y) - \psi(\alpha_u^{r^y})- \beta_u^{r^y}\\
		\frac{1}{\beta_u^{r^y}} \sum\nolimits_{\substack{1 \leq t \leq T:u_t=u}} \Big(\frac{r_t^y}{\beta_u^{r^y}} -\alpha_u^{r^y} \Big)\\
		\sum\nolimits_{\substack{1 \leq t \leq T:u_t=u}} \ln(g_t^x) - \psi(\alpha_u^{g^x})- \beta_u^{g^x}\\
		\frac{1}{\beta_u^{g^x}} \sum\nolimits_{\substack{1 \leq t \leq T:u_t=u}} \Big(\frac{g_t^x}{\beta_u^{g^x}} -\alpha_u^{g^x} \Big)\\
		\sum\nolimits_{\substack{1 \leq t \leq T:u_t=u}} \ln(g_t^y) - \psi(\alpha_u^{g^y})- \beta_u^{g^y}\\
		\frac{1}{\beta_u^{g^y}} \sum\nolimits_{\substack{1 \leq t \leq T:u_t=u}} \Big(\frac{g_t^y}{\beta_u^{g^y}} -\alpha_u^{g^y} \Big)
		\end{pmatrix}.}
		\end{equation*}
\end{proposition}
The likelihood of the newly-introduced features follows Gamma distributions in analogy to the distributions of durations and amplitudes in the {\sl Markov model}; a proof of Proposition~\ref{prop:gradient_markov_dynamic} is given in Appendix~\ref{sec:Proof_Markov_Dynamic}\refextendedversion.
\fussy

\subsection{Fisher Kernel for the SceneWalk Model}
Given a scanpath $\bS = ((q_1, d_1),\ldots,(q_T,d_T))$ obtained on an image $\bX$, the gradient of the logarithmic likelihood under the SceneWalk model parameterized with $\btheta = (\zeta, c_F, \omega_A, \omega_F, \sigma_A, \sigma_F, \gamma, \lambda)$ is 
\begin{equation*}
\bg=\nabla_{\btheta}  \ln p(\bS|\bX,\btheta) = \nabla_{\btheta} \ln p(q_1|\btheta, \bX) + \sum_{t=2}^{T} \ln p(q_t|q_1,...,q_{t-1}, d_1,...,d_{t-1},\btheta, \bX).
\end{equation*}

\section{Empirical Study}
\label{sec:EmpiricalStudy}
This section explores the performance of the derived models and reference models for viewer identification. 

\subsection{Data Collection}
We track the eye movements of 32 participants between the ages of 18 and 49 as they view 106 images of natural scenes for 8 seconds per image; the participants' task is to memorize the images. 
Participants sit at a viewing distance of 60~cm to the monitor, with their heads positioned in a chin rest. %Participation is rewarded by either 8 Euros per hour or course credits. 
The monitor has a diagonal size of 61.4~cm, an aspect ratio of 16 by 10 (1920x1080~px), and a refresh rate of 100-120~Hz. The images are presented at a resolution of 1500$\times$1000~px, and therefore subtend 48 degree by 28 degree of visual angle. 
We record participants' eye movements using an Eyelink 1000 video-based, desktop-mounted eye tracker with a sampling rate of 1000~Hz monocularly using the participant’s dominant eye. All participants have normal or corrected-to-normal vision. 
%For calibration, we present a 9-point target grid, followed by a validation using the same procedure. 
%The eye tracker is re-calibrated during the experiment at least every 14 trials to ensure a consistently high quality of the data. 
%In order to reduce the central fixation bias; that is, the propensity to fixate the center of the monitor, we precede each image with a cross shown for 150~ms cross in a random position and ask participants to fixate the cross until it disappears~\citep{Rothkegel2017}. 
From the raw samples recorded by the eyetracker, we extract the scanpaths  using a velocity-based saccade detection algorithm  \citep{Engbert2003}.

\subsection{Reference Methods}
The natural reference methods for the {\sl Fisher SVMs} are the underlying generative methods {\sl Markov model}, {\sl Markov model with saccade dynamics}, and {\sl SceneWalk}. As an additional generative reference method, we use the model of \cite{Abdelwahab2016} which has no Fisher kernel because it is nonparametric.
Other prior work on biometric identification using eye movements varies with regard to the type of stimuli, and the features that are extracted from the scanpath. 
Stimuli used for viewer identification include viewing artificial stimuli~\citep{Bednarik2005,Kasprowski2004,Cuong2012,Juhola2013,Darwish2013,Yoon2014,Zhang2014,Srivastava2015,Rigas2016}, text documents~\citep{Bednarik2005,Holland2011,Rigas2016}, movies~\citep{Kinnunen2010} or images~\citep{Bednarik2005,Darwish2013}. 
Most approaches are designed to identify viewers on a specific stimulus, for example by applying graph matching techniques to the scanpaths produced on a specific face image \citep{Rigas2012}, or even by including a secondary identification task such as entering a PIN or password with the eye gaze \citep{Maeder2004,Kumar2007,DeLuca2008,Dunphy2008,Weaver2011,Cymek2014}.
Approaches that can be applied to novel stimuli at test time extract different kinds of fixational and saccadic features, such as fixation durations \citep{SilverBiggs2006,George2016} or saccade amplitudes \citep{George2016,Rigas2012,Juhola2013,Rigas2016},  velocities  \citep{Bednarik2005,SilverBiggs2006,Cuong2012,Rigas2012,Darwish2013,Juhola2013,Eberz2015,Rigas2016} and accelerations \citep{Rigas2012,Darwish2013,Eberz2015,Rigas2016}, and either aggregate these over the whole scanpath \citep{SilverBiggs2006,Holland2011,Kinnunen2010,Darwish2013,George2016}, or compute the similarity of scanpaths by applying statistical tests to the distributions of the extracted  features \citep{Holland2013,Rigas2016}. 
As reference methods, we only consider methods that allow different stimuli for training and testing. 
As representative aggregational reference method, we choose the model by {\sl Holland and Komogortsev (2011)}\nocite{Holland2011}. 
As statistal reference approaches we use the seminal {\sl CEM-B} method~\citep{Holland2013} and the current state-of-the-art model  {\sl CEM-B with saccade dynamics}~\citep{Rigas2016}.

%Other existing methods that allow different stimuli for training and testing can be classified into approaches which  aggregate the extracted 
%%saccadic and fixational 
%features over the relevant recording window and statistical approaches that compute the similarity of scanpaths by applying statistical tests to the distributions of the extracted %saccadic and fixational features
%As representative aggregational reference method, we choose the model by {\sl Holland and Komogortsev (2011)}~\citep{Holland2011}. 
%As statistal reference approaches we use the seminal {\sl CEM-B} method~\citep{Holland2013} and the current state-of-the-art model by  {\sl Rigas et al. (2016)}~\citep{Rigas2016}.

\subsection{Evaluation Setting}
Each subject views a random subset of 106 photographs out of 376 photographs. We split the data into 50\% training and 50\% test data along photographs per subject, such that no photograph appears in both the training and test data. We average the identification accuracy across 5 random splits and study it as a function of the number of images seen at test time. All hyper-parameters of all methods are tuned by grid search using 3-fold cross validation on the training portion of the data.

\subsection{Results}
\subsubsection{Identification Accuracy}
\begin{figure}[t!]
	\centering 
	\includegraphics[clip, trim=1.0cm 0.0cm 6.7cm 0.0cm, width=0.8\textwidth]{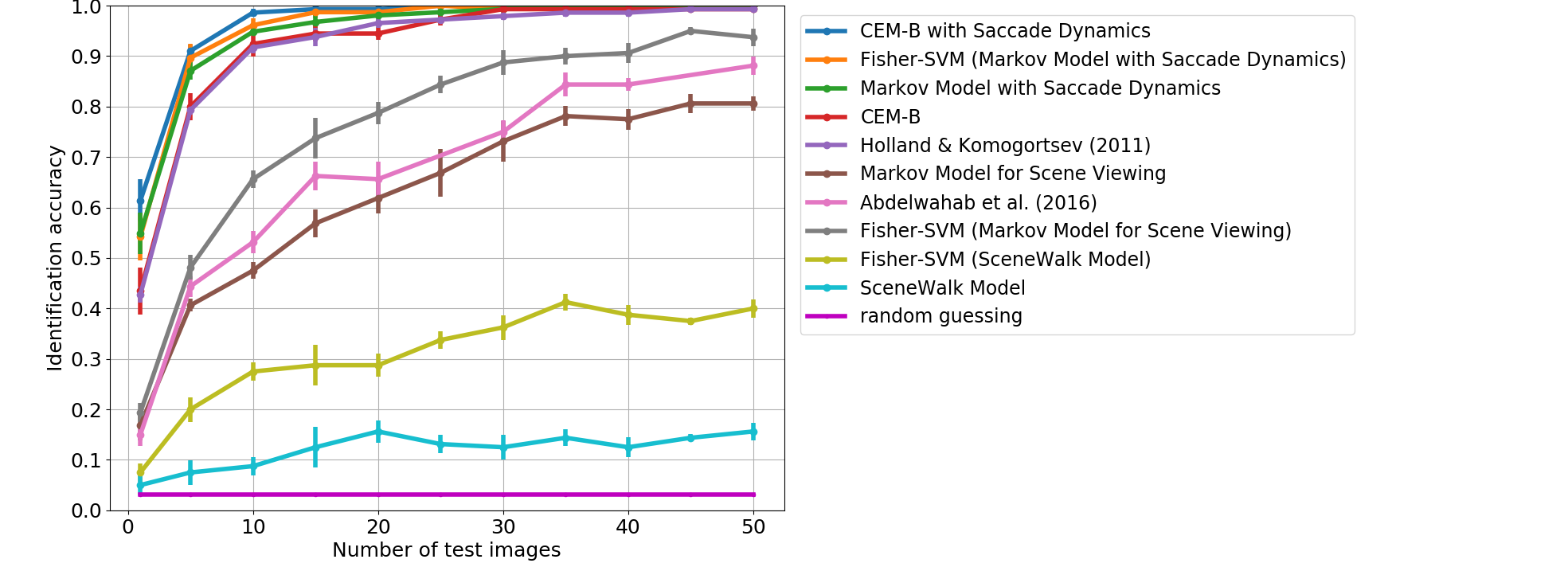}
	\caption{Identification accuracy (32 subjects) of all compared models as a function of the number of images seen at test time. Error bars show the standard error. Training was performed with 50 images per subject.}.
	\label{fig:identification accuracy all }
\end{figure}

Figure~\ref{fig:identification accuracy all } compares the identification accuracy as a function of the number of images that have been viewed at test time. The {\sl SceneWalk} model achieves the lowest identification accuracy; we attribute this to the fact that the classification can only be based on the saccade amplitudes and fixation durations since no other aspect of the scanpath is described by {\sl SceneWalk}. The {\sl Fisher SVM} for the {\sl SceneWalk} model improves the classification accuracy dramatically ($p<0.01$ for more than one test image).
The {\sl Markov model} has the second-lowest performance; in addition to saccade amplitudes and fixation durations it also models saccade durations and directions. Again, the {\sl Fisher SVM} on the {\sl Markov model} model improves the identification accuracy significantly over the generative model itself.
The non-parametric model of {\sl Abdelwahab et al.~(2016)} outperforms the {\sl Markov model} but is outperformed by the {\sl Fisher SVM} based on the {\sl Markov model}.
The {\sl Markov model with saccade dynamics} is the best-performing generative model. The performance comparison between this generative model and the {\sl Fisher SVM} based on it  is consistent with our previous observation that the {\sl Fisher SVM} improves the classification accuracy of the underlying generative model; but in this case, the differences are not statistically significant.
The {\sl CEM-B with saccade dynamics} performs comparably to the {\sl Fisher SVM} based on the {\sl Markov model with saccade dynamics}; differences are not significant. {\sl CEM-B with saccade dynamics} uses the largest feature set; in addition to the features extracted by the {\sl Markov model with saccade dynamics}, it extracts the saccadic peak velocity, absolute starting times of fixations and saccades, and the fixation locations on the screen.  

\subsubsection{Execution Time of Identification}
\begin{figure}[t!]
	\centering 
	\includegraphics[clip, trim=0.0cm 0.0cm 2.0cm 2.0cm, width=0.4\textwidth]{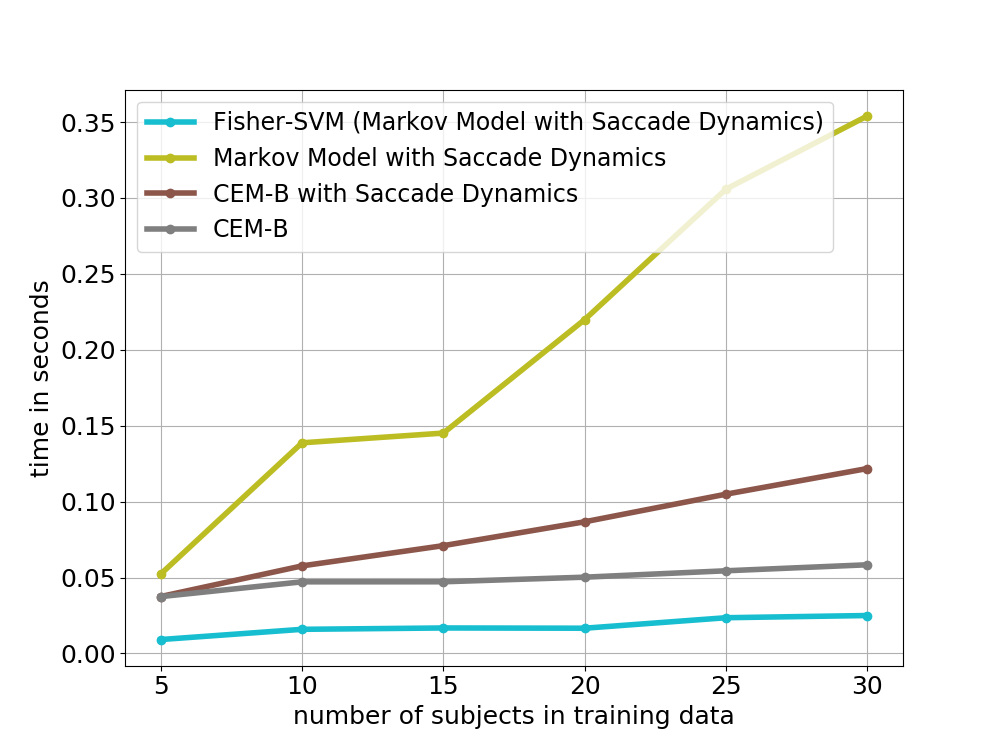}
	\caption{Execution time in seconds to identify one subject, after viewing one single image, as a function of the number of subjects in training data. %Execution was performed on a single two-core CPU (Intel Core i7-6600U, 2.6GHz).
	}
	\label{fig:identification time}
\end{figure}

We compare the execution time of the classification models for eye-movements from viewing one image and study it as a function of the number of persons in the training data. Figure~\ref{fig:identification time} show execution times on a single two-core CPU (Intel Core i7-6600U, 2.6GHz). The {\sl Fisher SVM} (based on any generative model) is a generalized linear multi-class classifier; it has the lowest execution time, and the slope of the execution time over the number of persons (classes) is the lowest. The {\sl CEM-B} method has a similar gradient but a higher absolute execution time. {\sl CEM-B with saccade dynamics} extracts a larger set of distributional features and compares these features to the profiles of each user. The {\sl Markov model with saccade dynamics} has to infer the likelihood of the observation sequence under each user-specific model and is therefore the slowest model by comparison.

\section{Conclusion}
\label{sec:Conclusion}
We have adapted a generative model for eye gaze during reading~\citep{Landwehr2014} to scene viewing. We have integrated features that describe the saccade dynamics into this {\sl Markov model with saccade dynamics}. Starting from these models and the known generative {\sl SceneWalk} model for eye gaze during scene viewing, we have derived Fisher kernels for discriminative classification. Whereas generative models are trained to maximize the regularized likelihood of the observed gaze sequences, a {\sl Fisher SVM} based on these generative models directly maximizes the classifier's ability to identify viewers based on their eye gaze. Experimentally, we find that the {\sl Fisher SVM} generally improves identification accuracy compared to the underlying generative model. %This improvement is particularly substantial in models with comparably low base accuracy. 
In terms of identification accuracy, the {\sl Fisher SVM with saccade dynamics} performs comparably to {\sl CEM-B with saccade dynamics} which extracts a larger set of distributional features from the scanpath; in terms of execution time, the {\sl Fisher SVM} with any generative model of eye gaze is the fastest method in our comparison. We conclude that while Fisher SVMs improve the identification accuracy compared to the underlying generative model, the selection of features that are described by the generative model are crucial. 

\section*{Acknowledgments}
This work was partially funded by the German Science Foundation under grant SFB 1294 (project number 318763901).
\appendix
\section{Appendix}

\subsection{Proof of Proposition~\ref{prop:gradient_markov}}
\label{sec:Proof_Markov}
\begin{proof}[Proposition~\ref{prop:gradient_markov}]
	The likelihood of the Markov Model for Scene Viewing factorizes as
	\begin{equation*}
	\ln p(\bS|\bX,\btheta) = \sum_{t=1}^{T} \ln \mathrm{Mult}(u_t| \bpi) + \sum_{t=1}^{T} \ln p(a_t | u_t,\balpha^a,\bbeta^a) 
	+ \sum_{t=1}^{T} \ln p(d_t | u_t, \balpha^d,\bbeta^d). 
	\end{equation*}
	For the multinomial distribution, 
	\begin{equation*}
	\sum\nolimits_{t=1}^{T} \ln \mathrm{Mult}(u_t| \bpi) = \ln \frac{T!}{\prod\nolimits_{u=1}^{4} K_u!} + \sum\nolimits_{u=1}^4 K_u \ln \pi_u
	\end{equation*}
	and thus for $u \in \{1,2,3,4\}$, we have that 
	%\begin{equation*}
	$\frac{\partial \ln p(\bS|\bX,\btheta)}{ \partial \pi_u} = \frac{K_u}{\pi_u}.$
	%\end{equation*}
	The likelihoods for the saccade amplitudes and for the fixation durations are defined analogously. Therefore, we only derive the gradient of the likelihood for a sequence of amplitudes. 
	As discussed in Section~\ref{sec:AdaptedLandwehr} (Equation~\ref{eq:likelihood_factorized_landwehr}), the likelihood of saccade amplitudes (and fixation durations) 
	factorizes over the different saccade types $u$.
	Its partial  derivative with respect to $\alpha_u^a$ therefore is
	\begin{align}
	\MoveEqLeft \frac{\partial} {\partial \alpha_u^a} \sum_{1 \leq t \leq T:u_t=u} \ln p(a_t | u_t,\balpha^a,\bbeta^a) \notag\\
	&= \frac{\partial} {\partial \alpha_u^a}  \sum_{1 \leq t \leq T:u_t=u} \ln \mathcal{G}(a_t| \alpha_u^a, \beta_u^a)\notag\\
	& = \frac{\partial }{\partial \alpha_u^a}\sum_{1 \leq t \leq T:u_t=u} 
	\ln \left(a_t^{\alpha_u^a-1}\operatorname{exp}({-\frac{a_t}{\beta_u^a})}\right) \notag
	-\ln(\Gamma(\alpha_u^a){(\beta_u^a)}^{\alpha_u^a})\notag\\
	&=\sum\limits_{\substack{1 \leq t \leq T:u_t=u}} \ln(a_t) - \psi(\alpha_u^a)- \beta_u^a  \label{eq:gradalpha}
	\end{align}
	and its partial derivative with respect to $\beta_u^a$ is  
	
	\begin{align}
	\MoveEqLeft\frac{\partial} {\partial \beta_u^a} \sum_{1 \leq t \leq T:u_t=u} \ln p(a_t | u_t, \alpha^a,\beta^a) \notag\\
	& = \frac{\partial} {\partial \beta_{u,m}^a} \sum_{1 \leq t \leq T:u_t=u} \ln \mathcal{G}(a_t| \alpha_u^a, \beta_u^a) \notag\\
	& = \frac{\partial }{\partial \beta_{u,m}^a} \sum_{1 \leq t \leq T:u_t=u} 
	\ln \left(a_t^{\alpha_u^a-1}\operatorname{exp}({-\frac{a_t}{\beta_u^a})}\right) \notag
	-\ln(\Gamma(\alpha_u^a){(\beta_u^a)}^{\alpha_u^a})\\
	&=\frac{1}{\beta_u^a} \sum\limits_{\substack{1 \leq t \leq T:u_t=u}} \Big(\frac{a_t}{\beta_u^a} -\alpha_u^a \Big).\label{eq:gradbeta}
	\end{align}
	In Equation~\ref{eq:gradalpha}, we exploit that the derivative of the log-gamma function is given by the digamma function $\psi$---\ie $\frac{d}{dx}\ln \Gamma(x)=\psi(x)$. The claim then follows from straightforward calculation.
\end{proof}

\subsection{Likelihood of the Markov Model with Saccade Dynamics}\label{ap:xlandwehr}
We maximize the likelihood 
\begin{equation}
\label{eq:maximum_likelihood_saccade_dynamics}
\btheta^* = \argmax_{\btheta} \sum_{i=1}^k \ln p(\bar\bS_i|\bar\bX_i,\btheta).
\end{equation}
With the saccade dynamics, the likelihood of Equation~\ref{eq:likelihood_factorized_landwehr} updates to
\begin{multline}
\btheta^* = \argmax_{\bpi,\balpha,\bbeta,\bgamma,\bdelta} \bigg(\sum_{i=1}^k \sum_{t=1}^{T_i} \ln \mathrm{Mult}(u^{(i)}_t| \bpi) 
+ \sum_{i=1}^k \sum_{t=1}^{T_i} \ln p(a^{(i)}_t | u^{(i)}_t,\balpha^a,\bbeta^a)  \\
+ \sum_{i=1}^k \sum_{t=1}^{T_i} \ln p(d^{(i)}_t | u^{(i)}_t,\balpha^d,\bbeta^d) 
+ \sum_{i=1}^k \sum_{t=1}^{T_i} \ln p(v^{(i)}_t | u^{(i)}_t,\balpha^v,\bbeta^v) \\
+ \sum_{i=1}^k \sum_{t=1}^{T_i} \ln p(w^{(i)}_t | u^{(i)}_t,\balpha^w,\bbeta^w)
+ \sum_{i=1}^k \sum_{t=1}^{T_i} \ln p(r^{(i,x)}_t | u^{(i)}_t,\balpha^{r^x},\bbeta^{r^x}) \\
+ \sum_{i=1}^k \sum_{t=1}^{T_i} \ln p(r^{(i,y)}_t | u^{(i)}_t,\balpha^{r^y},\bbeta^{r^y}) 
+ \sum_{i=1}^k \sum_{t=1}^{T_i} \ln p(g^{(i,x)}_t | u^{(i)}_t,\balpha^{g^x},\bbeta^{g^x}) \\
+ \sum_{i=1}^k \sum_{t=1}^{T_i} \ln p(g^{(i,y)}_t | u^{(i)}_t,\balpha^{g^y},\bbeta^{g^y})\bigg) 
\label{eq:likelihood_factorized_saccade_dynamics}
\end{multline}

\subsection{Proof of Proposition~\ref{prop:gradient_markov_dynamic}}
\label{sec:Proof_Markov_Dynamic}
\begin{proof}[Proposition 2]
	The likelihood of the {\sl Markov Model with Saccade Dynamics}  factorizes as
	\begin{multline}
	\ln p(\bS|\bX,\btheta) = \bigg(\sum_{t=1}^{T} \ln \mathrm{Mult}(u_t| \bpi) 
	+ \sum_{t=1}^{T} \ln p(a_t | u_t,\balpha^a,\bbeta^a)  \\
	+ \sum_{t=1}^{T} \ln p(d_t | u_t,\balpha^d,\bbeta^d) 
	+ \sum_{t=1}^{T} \ln p(v_t | u_t,\balpha^v,\bbeta^v) \\
	+ \sum_{t=1}^{T} \ln p(w_t | u_t,\balpha^w,\bbeta^w)
	+ \sum_{t=1}^{T} \ln p(r_t | u_t,\balpha^{r^x},\bbeta^{r^x}) \\
	+ \sum_{t=1}^{T} \ln p(r_t | u_t,\balpha^{r^y},\bbeta^{r^y}) 
	+ \sum_{t=1}^{T} \ln p(g^{(x)}_t | u_t,\balpha^{g^x},\bbeta^{g^x}) \\
	+ \sum_{t=1}^{T} \ln p(g^{(y)}_t | u_t,\balpha^{g^y},\bbeta^{g^y})\bigg) 
	\label{eq:likelihood_factorized_saccade_dynamics}
	\end{multline}
	In analogy to the likelihood defined for the {\sl Markov Model for Scene Viewing},  for the multinomial distribution of the saccade types we have  
	\begin{equation}
	\sum\nolimits_{t=1}^{T} \ln \mathrm{Mult}(u_t| \bpi) = \ln \frac{T!}{\prod\nolimits_{u=1}^{4} K_u!} + \sum\nolimits_{u=1}^4 K_u \ln \pi_u,
	\end{equation}
	and thus for $u \in \{1,2,3,4\}$, we have 
	\begin{equation}
	\frac{\partial \ln p(\bS|\bX,\btheta)}{ \partial \pi_u} = \frac{K_u}{\pi_u}.
	\end{equation}
	Since the likelihoods of all remaining features (i.e. fixation durations $d_1,...,d_T$, saccade mean velocities $v_1,...,v_T$, saccade mean accelerations $w_1,...,w_T$, horizontal $r_1^x,...,r_T^x$ and vertical $r_1^y,...,r_T^y$ peak-acceleration-to-deceleration ratios, horizontal $g_1^x,...,g_T^x$ and vertical $g_1^y,...,g_T^y$ saccade vigors) are analogous, we only derive the gradient of the amplitude likelihood. 
	The likelihood of saccade amplitudes further 
	factorizes over the different saccade types $u$.
	The partial  derivative of the logarithmic likelihood with respect to $\alpha_u^a$ therefore is
	\begin{align}
	\MoveEqLeft \frac{\partial} {\partial \alpha_u^a} \sum_{1 \leq t \leq T:u_t=u} \ln p(a_t | u_t,\balpha^a,\bbeta^a) \notag\\
	&= \frac{\partial} {\partial \alpha_u^a}  \sum_{1 \leq t \leq T:u_t=u} \ln \mathcal{G}(a_t| \alpha_u^a, \beta_u^a)\notag\\
	& = \frac{\partial }{\partial \alpha_u^a}\sum_{1 \leq t \leq T:u_t=u} 
	\ln \left(a_t^{\alpha_u^a-1}\operatorname{exp}({-\frac{a_t}{\beta_u^a})}\right) \notag
	-\ln(\Gamma(\alpha_u^a){(\beta_u^a)}^{\alpha_u^a})\notag\\
	&=\sum\limits_{\substack{1 \leq t \leq T:u_t=u}} \ln(a_t) - \psi(\alpha_u^a)- \beta_u^a  \label{eq:gradalpha}
	\end{align}
	
	and the partial derivative with respect to $\beta_u^a$ is  
	\begin{align}
	\MoveEqLeft\frac{\partial} {\partial \beta_u^a} \sum_{1 \leq t \leq T:u_t=u} \ln p(a_t | u_t, \alpha^a,\beta^a) \notag\\
	& = \frac{\partial} {\partial \beta_{u,m}^a} \sum_{1 \leq t \leq T:u_t=u} \ln \mathcal{G}(a_t| \alpha_u^a, \beta_u^a) \notag\\
	& = \frac{\partial }{\partial \beta_{u,m}^a} \sum_{1 \leq t \leq T:u_t=u} 
	\ln \left(a_t^{\alpha_u^a-1}\operatorname{exp}({-\frac{a_t}{\beta_u^a})}\right) \notag
	-\ln(\Gamma(\alpha_u^a){(\beta_u^a)}^{\alpha_u^a})\\
	&=\frac{1}{\beta_u^a} \sum\limits_{\substack{1 \leq t \leq T:u_t=u}} \Big(\frac{a_t}{\beta_u^a} -\alpha_u^a \Big).\label{eq:gradbeta}
	\end{align}
	In Equation~\ref{eq:gradalpha}, we exploit that the derivative of the log-gamma function is given by the digamma function $\psi$---\ie $\frac{d}{dx}\ln \Gamma(x)=\psi(x)$. The claim now follows from straightforward calculation.
\end{proof}

\subsection{SceneWalk Log-Likelihood Gradient}
The likelihood of a scanpath $\bS$ in the {\sl SceneWalk} model  factorizes as
\begin{equation}
\ln p(\bS|\bX,\btheta) = \sum_{t=1}^{T} \ln p(q_{t+1}|q_1,...,q_{t}, d_1,...,d_{t},\btheta),
\end{equation} 
The {\sl Scene Walk} model consists of eight parameters: $\btheta = (\zeta, c_F, \lambda, \gamma, \omega_A, \omega_F, \sigma_A, \sigma_F)$.
We use the model definition from Section~\ref{sec:SceneWalk} to derive the partial derivatives with respect to each of the model parameters in order to obtain the gradient $\bg$ of the logarithmic likelihood:
\begin{equation}
\bg=\nabla_{\btheta}  \ln p(\bS|\bX,\btheta) = \nabla_{\btheta} \sum_{t=1}^{T} \ln p(q_{t+1}|q_1,...,q_{t}, d_1,...,d_{t},\btheta, \bX)
\end{equation} 
To reduce notational clutter, let $\pi_{q_{t+1}} = p(q_{t+1}|q_1,...,q_{t}, d_1,...,d_{t}, \btheta,\bX)$.

\subsubsection{Partial derivative with respect to parameter $\zeta$}
\begin{align}
\frac{\partial} {\partial \zeta} \sum_{1 \leq t \leq T} \ln \pi_{q_{t+1}}
&= \frac{\partial} {\partial \zeta}  \sum_{1 \leq t \leq T} \ln \left( (1-\zeta) \frac{\mathbf{U}_{t}(i_{t+1}, j_{t+1})}{\sum_{i,j} \mathbf{U}_{t}(i, j)} + \zeta \frac{1}{\sum_{i,j} 1} \right)\notag\\
&= \sum_{1 \leq t \leq T} \frac{1}{\pi_{q_{t+1}}} \left(- \frac{\mathbf{U}_{t}(i_{t+1}, j_{t+1})}{\sum_{i,j} \mathbf{U}_{t}(i, j)} + \frac{1}{\sum_{i,j} 1}\right)    
\label{eq:gradzeta}
\end{align}

\subsubsection{Partial derivative with respect to parameter $c_F$}
\begin{align}
\MoveEqLeft \frac{\partial} {\partial c_F} \sum_{1 \leq t \leq T} \ln \pi_{q_{t+1}} \notag\\
&= \sum_{1 \leq t \leq T}\frac{1}{\pi_{q_{t+1}}} \frac{\partial \pi_{q_{t+1}}}{\partial c_F}\label{eq:gradcf}\\
&= \sum_{1 \leq t \leq T} \frac{1}{\pi_{q_{t+1}}} (1-\zeta) \frac{\frac{\partial \mathbf{U}_{t}(i_{t+1}, j_{t+1})}{\partial c_F} \sum_{i,j} \mathbf{U}_{t}(i, j) - \mathbf{U}_{t}(i_{t+1}, j_{t+1}) \sum_{i,j} \frac{\partial \mathbf{U}_{t}(i, j)}{\partial c_F}}{(\sum_{i,j} \mathbf{U}_{t}(i, j))^2}  
\notag
\end{align}

The partial derivative of the combined potential  $\mathbf{U}_{t}(i_{t+1}, j_{t+1})$ for parameter $c_F$ is 
\begin{equation}
\frac{\partial \mathbf{U}_{t}(i_{t+1}, j_{t+1})}{\partial c_F} = - \frac{\mathbf{F}_{t}(i_{t+1}, j_{t+1})^\gamma}{\sum_{i,j} \mathbf{F}_{t}(i, j)^\gamma}
\end{equation}

\subsubsection{Partial derivative with respect to parameter $\lambda$}
\begin{align}
\MoveEqLeft \frac{\partial} {\partial \lambda} \sum_{1 \leq t \leq T} \ln \pi_{q_{t+1}} \notag\\
&= \sum_{1 \leq t \leq T} \frac{1}{\pi_{q_{t+1}}} \frac{\partial \pi_{q_{t+1}}}{\partial \lambda}\label{eq:gradlambda}\\
&= \sum_{1 \leq t \leq T} \frac{1}{\pi_{q_{t+1}}} (1-\zeta) \frac{\frac{\partial \mathbf{U}_{t}(i_{t+1}, j_{t+1})}{\partial \lambda} \sum_{i,j} \mathbf{U}_{t}(i, j) - \mathbf{U}_{t}(i_{t+1}, j_{t+1}) \sum_{i,j} \frac{\partial \mathbf{U}_{t}(i, j)}{\partial \lambda}}{(\sum_{i,j} \mathbf{U}_{t}(i, j))^2} \notag 
\end{align}
And the partial derivative of the combined potential $ \mathbf{U}_{t}(i_{t+1}, j_{t+1})$ for parameter $\lambda$ is
\begin{align}
\MoveEqLeft \frac{\partial \mathbf{U}_{t}(i_{t+1}, j_{t+1})}{\partial \lambda} \label{eq:gradUlambda}\\
&= \frac{\log \mathbf{A}_{t}(i_{t+1}, j_{t+1}) {\mathbf{A}_{t}(i_{t+1}, j_{t+1})}^{\lambda}}{\sum_{i,j} \mathbf{A}_{t}(i, j)^{\lambda}}
- \frac{ {\mathbf{A}_{t}(i_{t+1}, j_{t+1})}^{\lambda}}{(\sum_{i,j} \mathbf{A}_{t}(i, j)^{\lambda})^2} \sum_{i,j} \log \mathbf{A}_{t}(i, j) {\mathbf{A}_{t}(i, j)}^{\lambda}. \notag
\end{align}

\subsubsection{Partial derivative with respect to parameter $\gamma$}
\begin{align}
\MoveEqLeft \frac{\partial} {\partial \gamma} \sum_{1 \leq t \leq T} \ln \pi_{q_{t+1}} \notag\\
&= \sum_{1 \leq t \leq T} \frac{1}{\pi_{q_{t+1}}} \frac{\partial \pi_{q_{t+1}}}{\partial \gamma}\label{eq:gradgamma}\\
&= \sum_{1 \leq t \leq T} \frac{1}{\pi_{q_{t+1}}} (1-\zeta) \frac{\frac{\partial \mathbf{U}_{t}(i_{t+1}, j_{t+1})}{\partial \gamma} \sum_{i,j} \mathbf{U}_{t}(i, j) - \mathbf{U}_{t}(i_{t+1}, j_{t+1}) \sum_{i,j} \frac{\partial \mathbf{U}_{t}(i, j)}{\partial \gamma}}{(\sum_{i,j} \mathbf{U}_{t}(i, j))^2}, \notag 
\end{align}
where the partial derivative of the combined potential $ \mathbf{U}_{t}(i_{t+1}, j_{t+1})$ for parameter $\gamma$ is
\begin{align}
\MoveEqLeft \frac{\partial \mathbf{U}_{t}(i_{t+1}, j_{t+1})}{\partial \gamma} \label{eq:gradUgamma}
\\
&= -c_F \Big(\frac{\log \mathbf{F}_{t}(i_{t+1}, j_{t+1}) {\mathbf{F}_{t}(i_{t+1}, j_{t+1})}^{\gamma}}{\sum_{i,j} {\mathbf{F}_{t}(i, j)}^{\gamma}} 
- \frac{ {\mathbf{F}_{t}(i_{t+1}, j_{t+1})}^{\gamma}}{(\sum_{i,j} {\mathbf{F}_{t}(i, j)}^{\gamma} )^2} \sum_{i,j} \log \mathbf{F}_{t}(i, j) {\mathbf{F}_{t}(i, j)}^{\gamma} \Big).\notag
\end{align}

\subsubsection{Partial derivative with respect to parameter $\omega_A$}
\begin{align}
\MoveEqLeft \frac{\partial} {\partial \omega_A} \sum_{1 \leq t \leq T} \ln \pi_{q_{t+1}} \notag\\
&= \sum_{1 \leq t \leq T} \frac{1}{\pi_{q_{t+1}}} \frac{\partial \pi_{q_{t+1}}}{\partial \omega_A}\label{eq:gradomegaA}\\
&= \sum_{1 \leq t \leq T} \frac{1}{\pi_{q_{t+1}}} (1-\zeta) \frac{\frac{\partial \mathbf{U}_{t}(i_{t+1}, j_{t+1})}{\partial \omega_A} \sum_{i,j} \mathbf{U}_{t}(i, j) - \mathbf{U}_{t}(i_{t+1}, j_{t+1}) \sum_{i,j} \frac{\partial \mathbf{U}_{t}(i, j)}{\partial \omega_A}}{(\sum_{i,j} \mathbf{U}_{t}(i, j))^2} \notag
\end{align}
Here, the partial derivative of the combined potential $ \mathbf{U}_{t}(i_{t+1}, j_{t+1})$ for parameter $\omega_A$ is
\begin{align}
\MoveEqLeft \frac{\partial \mathbf{U}_{t}(i_{t+1}, j_{t+1})}{\partial \omega_A} \label{eq:gradUomegaA}\\
&= \frac{\frac{\partial {\mathbf{A}_{t}(i_{t+1}, j_{t+1})}^{\lambda}}{\partial \omega_A} \sum_{i,j} {\mathbf{A}_{t}(i, j)}^{\lambda} - {\mathbf{A}_{t}(i_{t+1}, j_{t+1})}^{\lambda} \sum_{i,j} \frac{\partial {\mathbf{A}_{t}(i, j)}^{\lambda}}{\partial \omega_A}}{(\sum_{i,j} {\mathbf{A}_{t}(i, j)}^{\lambda})^2}, \notag 
\end{align}
and the partial derivative of the attentional component $\mathbf{A}_{t}(i_{t+1}, j_{t+1})$ for parameter $\omega_A$ is
\begin{equation}
\frac{\partial \mathbf{A}_{t}(i_{t+1}, j_{t+1})}{\partial \omega_A} 
= \exp(- \omega_A d_t) \left(\frac{\partial \mathbf{A}_{t-1}}{\partial \omega_A} - d_t \left(\mathbf{A}_{t-1} - \frac{\mathbf{G}_{t}^A \mathbf{H}}{\sum_{i,j} \mathbf{G}_{t}^A(i,j) \mathbf{H}(i,j)}\right)\right). 
\label{eq:gradAomegaA}
\end{equation}

\subsubsection{Partial derivative with respect to parameter $\omega_F$}
\begin{align}
\MoveEqLeft \frac{\partial} {\partial \omega_F} \sum_{1 \leq t \leq T} \ln \pi_{q_{t+1}} \notag\\
&= \sum_{1 \leq t \leq T} \frac{1}{\pi_{q_{t+1}}} \frac{\partial \pi_{q_{t+1}}}{\partial \omega_F}\label{eq:gradomegaF}\\
&= \sum_{1 \leq t \leq T} \frac{1}{\pi_{q_{t+1}}} (1-\zeta) \frac{\frac{\partial \mathbf{U}_{t}(i_{t+1}, j_{t+1})}{\partial \omega_F} \sum_{i,j} \mathbf{U}_{t}(i, j) - \mathbf{U}_{t}(i_{t+1}, j_{t+1}) \sum_{i,j} \frac{\partial \mathbf{U}_{t}(i, j)}{\partial \omega_F}}{(\sum_{i,j} \mathbf{U}_{t}(i, j))^2}, \notag
\end{align}
where the partial derivative of the combined potential $ \mathbf{U}_{t}(i_{t+1}, j_{t+1})$ for parameter $\omega_F$ is
\begin{align}
\MoveEqLeft \frac{\partial \mathbf{U}_{t}(i_{t+1}, j_{t+1})}{\partial \omega_F} \label{eq:gradUomegaF}\\
&= -c_F \frac{\frac{\partial {\mathbf{F}_{t}(i_{t+1}, j_{t+1})}^{\gamma}}{\partial \omega_F} \sum_{i,j} {\mathbf{F}_{t}(i, j)}^{\gamma} - {\mathbf{F}_{t}(i_{t+1}, j_{t+1})}^{\gamma} \sum_{i,j} \frac{\partial {\mathbf{F}_{t}(i, j)}^{\gamma}}{\partial \omega_F}}{(\sum_{i,j} {\mathbf{F}_{t}(i, j)}^{\gamma})^2}, \notag
\end{align}
and the partial derivative of the inhibitory component $\mathbf{F}_{t}(i_{t+1}, j_{t+1})$ for parameter $\omega_F$ is
\begin{equation}
\frac{\partial \mathbf{F}_{t}(i_{t+1}, j_{t+1})}{\partial \omega_F} 
= \exp(- \omega_F d_t) \left(\frac{\partial \mathbf{F}_{t-1}}{\partial \omega_F} - d_t \left(\mathbf{F}_{t-1} - \frac{\mathbf{G}_{t}^F }{\sum_{i,j} \mathbf{G}_{t}^F(i,j)}\right)\right). \label{eq:gradFomegaF}
\end{equation}

\subsubsection{Partial derivative with respect to parameter $\sigma_A$}
\begin{align}
\MoveEqLeft \frac{\partial} {\partial \sigma_A} \sum_{1 \leq t \leq T} \ln \pi_{q_{t+1}} \notag\\
&= \sum_{1 \leq t \leq T} \frac{1}{\pi_{q_{t+1}}} \frac{\partial \pi_{q_{t+1}}}{\partial \sigma_A}\label{eq:gradsigmaA}\\
&= \sum_{1 \leq t \leq T} \frac{1}{\pi_{q_{t+1}}} (1-\zeta) \frac{\frac{\partial \mathbf{U}_{t}(i_{t+1}, j_{t+1})}{\partial \sigma_A} \sum_{i,j} \mathbf{U}_{t}(i, j) - \mathbf{U}_{t}(i_{t+1}, j_{t+1}) \sum_{i,j} \frac{\partial \mathbf{U}_{t}(i, j)}{\partial \sigma_A}}{(\sum_{i,j} \mathbf{U}_{t}(i, j))^2}. \notag 
\end{align}
The partial derivative of the combined potential $ \mathbf{U}_{t}(i_{t+1}, j_{t+1})$ for parameter $\sigma_A$ is 
\begin{align}
\MoveEqLeft \frac{\partial \mathbf{U}_{t}(i_{t+1}, j_{t+1})}{\partial \sigma_A}\label{eq:gradUsigmaA}\\
&= \frac{\frac{\partial {\mathbf{A}_{t}(i_{t+1}, j_{t+1})}^{\lambda}}{\partial \sigma_A} \sum_{i,j} {\mathbf{A}_{t}(i, j)}^{\lambda} - {\mathbf{A}_{t}(i_{t+1}, j_{t+1})}^{\lambda} \sum_{i,j} \frac{\partial {\mathbf{A}_{t}(i, j)}^{\lambda}}{\partial \sigma_A}}{(\sum_{i,j} {\mathbf{A}_{t}(i, j)}^{\lambda})^2} \notag 
\end{align}
and the partial derivative of the attention component $\mathbf{A}_{t}$ for $\sigma_A$ is
\begin{align}
\frac{\partial \mathbf{A}_{t}}{\partial \sigma_A} 
&= \frac{\partial \frac{\mathbf{G}_{t}^A \mathbf{S}}{\sum_{i,j} \mathbf{G}_{t}^A(i,j) \mathbf{S}(i,j)}}{\partial \sigma_A} + \exp(-\omega_A d_t) \left( \frac{\partial \mathbf{A}_{t-1}}{\partial \sigma_A} - \frac{\frac{\mathbf{G}_{t}^A \mathbf{S}}{\sum_{i,j} \mathbf{G}_{t}^A(i,j) \mathbf{S}(i,j)}}{\partial \sigma_A}\right) 
\label{eq:gradAsigmaA}
\end{align}
and the partial derivative of the Gaussian attention for $\sigma_A$ is 
\begin{align}
\frac{\partial \mathbf{G}_{t}^A \mathbf{S}}{\partial \sigma_A}
=& \frac{\partial \mathbf{G}_{t}^A(i_{t+1}, j_{t+1}, i_t, j_t) \mathbf{S}(i_{t+1}, j_{t+1})}{\partial \sigma_A} \notag\\
=& \mathbf{S}(i_{t+1}, j_{t+1}) \Biggl(\exp\left(- \frac{(i_{t+1}-i_t)^2 + (j_{t+1}-j_t)^2}{2 \sigma_A^2}\right) \label{eq:gradGsigmaA}\\
&\left( \frac{1}{2 \pi \sigma_A^2} \frac{(i_{t+1}-i_t)^2 + (j_{t+1}-j_t)^2}{\sigma_A^3} - \frac{1}{\pi \sigma_A^3}\right)   \Biggr) 
\notag
\end{align}
Where $(i_{t+1}, j_{t+1})$ are the position of fixation $q_{t+1}$ and $(i_{t}, j_{t})$ of the previous fixation $q_t$.

\subsubsection{Partial derivative with respect to parameter $\sigma_F$}
\begin{align}
\MoveEqLeft \frac{\partial} {\partial \sigma_F} \sum_{1 \leq t \leq T} \ln \pi_{q_{t+1}} \notag\\
&= \sum_{1 \leq t \leq T} \frac{1}{\pi_{q_{t+1}}} \frac{\partial \pi_{q_{t+1}}}{\partial \sigma_F}
\label{eq:gradsigmaF}\\
&= \sum_{1 \leq t \leq T} \frac{1}{\pi_{q_{t+1}}} (1-\zeta) \frac{\frac{\partial \mathbf{U}_{t}(i_{t+1}, j_{t+1})}{\partial \sigma_F} \sum_{i,j} \mathbf{U}_{t}(i, j) - \mathbf{U}_{t}(i_{t+1}, j_{t+1}) \sum_{i,j} \frac{\partial \mathbf{U}_{t}(i, j)}{\partial \sigma_F}}{(\sum_{i,j} \mathbf{U}_{t}(i, j))^2} \notag 
\end{align}
where the partial derivative of the potential for $\sigma_F$ is
\begin{align}
\MoveEqLeft \frac{\partial \mathbf{U}_{t}(i_{t+1}, j_{t+1})}{\partial \sigma_F} \label{eq:gradUsigmaF}
\\
&= -c_F \frac{\frac{\partial {\mathbf{F}_{t}(i_{t+1}, j_{t+1})}^{\gamma}}{\partial \sigma_F} \sum_{i,j} {\mathbf{F}_{t}(i, j)}^{\gamma} - {\mathbf{F}_{t}(i_{t+1}, j_{t+1})}^{\gamma} \sum_{i,j} \frac{\partial {\mathbf{F}_{t}(i, j)}^{\gamma}}{\partial \sigma_F}}{(\sum_{i,j} {\mathbf{F}_{t}(i, j)}^{\gamma})^2} \notag
\end{align}
and the partial derivative of the inhibition component $\mathbf{F}_{t}$ for $\sigma_F$ is
\begin{equation}
\frac{\partial \mathbf{F}_{t}}{\partial \sigma_F} 
= \frac{\partial \frac{\mathbf{G}_{t}^F}{\sum_{i,j} \mathbf{G}_{t}^F(i,j)}}{\partial \sigma_F} + \exp(-\omega_F d_t) \left( \frac{\partial \mathbf{F}_{t-1}}{\partial \sigma_F} - \frac{\partial \frac{\mathbf{G}_{t}^F}{\sum_{i,j} \mathbf{G}_{t}^F(i,j) }}{\partial \sigma_F}\right)
\label{eq:gradFsigmaF}
\end{equation}
and the partial derivation of the Gaussian inhibition component $\mathbf{G}_{t}^F$ for $\sigma_F$ is
\begin{align}
\frac{\partial \mathbf{G}_{t}^F}{\partial \sigma_F} 
&= \frac{\partial \mathbf{G}_{t}^F(i_{t+1}, j_{t+1}, i_t, j_t)}{\partial \sigma_F} \notag\\
&= \exp(- \frac{(i_{t+1}-i_t)^2 + (j_{t+1}-j_t)^2}{2 \sigma_F^2}) \label{eq:gradGsigmaF}\\
&\left( \frac{1}{2 \pi \sigma_F^2} \frac{(i_{t+1}-i_t)^2 + (j_{t+1}-j_t)^2}{\sigma_F^3} - \frac{1}{\pi \sigma_F^3}\right)\notag
\end{align}
Where $(i_{t+1}, j_{t+1})$ is the location of fixation $q_{t+1}$ and $(i_{t}, j_{t})$ the location of the previous fixation $q_t$.

\bibliographystyle{elsarticle-harv}
\bibliography{procs_kes_makowski}

\begin{thebibliography}{35}
\expandafter\ifx\csname natexlab\endcsname\relax\def\natexlab#1{#1}\fi
\providecommand{\url}[1]{\texttt{#1}}
\providecommand{\href}[2]{#2}
\providecommand{\path}[1]{#1}
\providecommand{\DOIprefix}{doi:}
\providecommand{\ArXivprefix}{arXiv:}
\providecommand{\URLprefix}{URL: }
\providecommand{\Pubmedprefix}{pmid:}
\providecommand{\doi}[1]{\href{http://dx.doi.org/#1}{\path{#1}}}
\providecommand{\Pubmed}[1]{\href{pmid:#1}{\path{#1}}}
\providecommand{\bibinfo}[2]{#2}
\ifx\xfnm\relax \def\xfnm[#1]{\unskip,\space#1}\fi
%Type = Inproceedings
\bibitem[{Abdelwahab et~al.(2016)Abdelwahab, Kliegl and
  Landwehr}]{Abdelwahab2016}
\bibinfo{author}{Abdelwahab, A.}, \bibinfo{author}{Kliegl, R.},
  \bibinfo{author}{Landwehr, N.}, \bibinfo{year}{2016}.
\newblock \bibinfo{title}{A semiparametric model for {B}ayesian reader
  identification}, in: \bibinfo{booktitle}{Proceedings of the 2016 Conference
  on Empirical Methods in Natural Language Processing}.
%Type = Book
\bibitem[{Baddeley et~al.(2015)Baddeley, Rubak and Turner}]{baddeley2015}
\bibinfo{author}{Baddeley, A.}, \bibinfo{author}{Rubak, E.},
  \bibinfo{author}{Turner, R.}, \bibinfo{year}{2015}.
\newblock \bibinfo{title}{Spatial point patterns: methodology and applications
  with R}.
\newblock \bibinfo{publisher}{Chapman and Hall/CRC}.
%Type = Article
\bibitem[{Baloh et~al.(1975)Baloh, Sills, Kumley and
  Honrubia}]{baloh1975quantitative}
\bibinfo{author}{Baloh, R.W.}, \bibinfo{author}{Sills, A.W.},
  \bibinfo{author}{Kumley, W.E.}, \bibinfo{author}{Honrubia, V.},
  \bibinfo{year}{1975}.
\newblock \bibinfo{title}{Quantitative measurement of saccade amplitude,
  duration, and velocity}.
\newblock \bibinfo{journal}{Neurology} \bibinfo{volume}{25},
  \bibinfo{pages}{1065--1065}.
%Type = Article
\bibitem[{Bargary et~al.(2017)Bargary, Bosten, Goodbourn, Lawrance-Owen, Hogg
  and Mollon}]{Bargary2017}
\bibinfo{author}{Bargary, G.}, \bibinfo{author}{Bosten, J.M.},
  \bibinfo{author}{Goodbourn, P.T.}, \bibinfo{author}{Lawrance-Owen, A.J.},
  \bibinfo{author}{Hogg, R.E.}, \bibinfo{author}{Mollon, J.},
  \bibinfo{year}{2017}.
\newblock \bibinfo{title}{Individual differences in human eye movements: {A}n
  oculomotor signature?}
\newblock \bibinfo{journal}{Vision Research} \bibinfo{volume}{141},
  \bibinfo{pages}{157--169}.
%Type = Inproceedings
\bibitem[{Bednarik et~al.(2005)Bednarik, Kinnunen, Mihaila and
  Fr\"anti}]{Bednarik2005}
\bibinfo{author}{Bednarik, R.}, \bibinfo{author}{Kinnunen, T.},
  \bibinfo{author}{Mihaila, A.}, \bibinfo{author}{Fr\"anti, P.},
  \bibinfo{year}{2005}.
\newblock \bibinfo{title}{Eye-movements as a biometric}, in:
  \bibinfo{booktitle}{Proceedings of the 14th Scandinavian Conference on Image
  Analysis ({SCIA} 2005)}, pp. \bibinfo{pages}{780--789}.
%Type = Inproceedings
\bibitem[{Cuong et~al.(2012)Cuong, Dinh and Ho}]{Cuong2012}
\bibinfo{author}{Cuong, N.}, \bibinfo{author}{Dinh, V.}, \bibinfo{author}{Ho,
  L.S.T.}, \bibinfo{year}{2012}.
\newblock \bibinfo{title}{Mel-frequency cepstral coefficients for eye movement
  identification}, in: \bibinfo{booktitle}{24th International Conference on
  Tools with Artificial Intelligence {(ICTAI)}}, pp. \bibinfo{pages}{253--260}.
%Type = Article
\bibitem[{Cymek et~al.(2014)Cymek, Venjakob, Ruff, Lutz, Hofmann and
  Roetting}]{Cymek2014}
\bibinfo{author}{Cymek, D.}, \bibinfo{author}{Venjakob, A.},
  \bibinfo{author}{Ruff, S.}, \bibinfo{author}{Lutz, O.M.},
  \bibinfo{author}{Hofmann, S.}, \bibinfo{author}{Roetting, M.},
  \bibinfo{year}{2014}.
\newblock \bibinfo{title}{Entering {PIN} codes by smooth pursuit eye
  movements}.
\newblock \bibinfo{journal}{Journal of Eye Movement Research}
  \bibinfo{volume}{7}, \bibinfo{pages}{1--11}.
%Type = Inproceedings
\bibitem[{Darwish and Pasquier(2013)}]{Darwish2013}
\bibinfo{author}{Darwish, A.}, \bibinfo{author}{Pasquier, M.},
  \bibinfo{year}{2013}.
\newblock \bibinfo{title}{Biometric identification using the dynamic features
  of the eyes}, in: \bibinfo{booktitle}{6th International Conference on
  Biometrics: {T}heory, Applications and Systems ({BTAS})}, pp.
  \bibinfo{pages}{1--6}.
%Type = Inproceedings
\bibitem[{{De Luca} et~al.(2007){De Luca}, Weiss, Hußmann and An}]{DeLuca2008}
\bibinfo{author}{{De Luca}, A.}, \bibinfo{author}{Weiss, R.},
  \bibinfo{author}{Hußmann, H.}, \bibinfo{author}{An, X.},
  \bibinfo{year}{2007}.
\newblock \bibinfo{title}{Eyepass -- eye-stroke authentication for public
  terminals}, in: \bibinfo{booktitle}{Extended Abstracts on Human Factors in
  Computing Systems {(CHI EA '08)}}, pp. \bibinfo{pages}{3003--3008}.
%Type = Inproceedings
\bibitem[{Dunphy et~al.(2008)Dunphy, Fitch and Olivier}]{Dunphy2008}
\bibinfo{author}{Dunphy, P.}, \bibinfo{author}{Fitch, A.},
  \bibinfo{author}{Olivier, P.}, \bibinfo{year}{2008}.
\newblock \bibinfo{title}{Gaze-contingent passwords at the {ATM}}, in:
  \bibinfo{booktitle}{4th Conference on Communication by Gaze Interaction
  {(COGAIN)}}, pp. \bibinfo{pages}{59--62}.
%Type = Inproceedings
\bibitem[{Eberz et~al.(2015)Eberz, Rasmussen, Lenders and
  Martinovic}]{Eberz2015}
\bibinfo{author}{Eberz, S.}, \bibinfo{author}{Rasmussen, K.},
  \bibinfo{author}{Lenders, V.}, \bibinfo{author}{Martinovic, I.},
  \bibinfo{year}{2015}.
\newblock \bibinfo{title}{Preventing lunchtime attacks: {F}ighting insider
  threats with eye movement biometrics}, in: \bibinfo{booktitle}{Network and
  Distributed System Security {(NDSS)} Symposium}.
%Type = Article
\bibitem[{Engbert and Kliegl(2003)}]{Engbert2003}
\bibinfo{author}{Engbert, R.}, \bibinfo{author}{Kliegl, R.},
  \bibinfo{year}{2003}.
\newblock \bibinfo{title}{Microsaccades uncover the orientation of covert
  attention}.
\newblock \bibinfo{journal}{Vision Research} \bibinfo{volume}{43},
  \bibinfo{pages}{1035--1045}.
%Type = Article
\bibitem[{Engbert et~al.(2015)Engbert, Trukenbrod, Barthelm{\'e} and
  Wichmann}]{engbert2015spatial}
\bibinfo{author}{Engbert, R.}, \bibinfo{author}{Trukenbrod, H.A.},
  \bibinfo{author}{Barthelm{\'e}, S.}, \bibinfo{author}{Wichmann, F.A.},
  \bibinfo{year}{2015}.
\newblock \bibinfo{title}{Spatial statistics and attentional dynamics in scene
  viewing}.
\newblock \bibinfo{journal}{Journal of Vision} \bibinfo{volume}{15},
  \bibinfo{pages}{14--14}.
%Type = Article
\bibitem[{George and Routray(2016)}]{George2016}
\bibinfo{author}{George, A.}, \bibinfo{author}{Routray, A.},
  \bibinfo{year}{2016}.
\newblock \bibinfo{title}{A score level fusion method for eye movement
  biometrics}.
\newblock \bibinfo{journal}{Pattern Recognition Letters} \bibinfo{volume}{82},
  \bibinfo{pages}{207--215}.
%Type = Incollection
\bibitem[{Henderson and Hollingworth(1998)}]{henderson1998eye}
\bibinfo{author}{Henderson, J.M.}, \bibinfo{author}{Hollingworth, A.},
  \bibinfo{year}{1998}.
\newblock \bibinfo{title}{Eye movements during scene viewing: An overview}, in:
  \bibinfo{booktitle}{Eye guidance in reading and scene perception}.
  \bibinfo{publisher}{Elsevier}, pp. \bibinfo{pages}{269--293}.
%Type = Inproceedings
\bibitem[{Holland and Komogortsev(2013)}]{Holland2013}
\bibinfo{author}{Holland, C.}, \bibinfo{author}{Komogortsev, O.},
  \bibinfo{year}{2013}.
\newblock \bibinfo{title}{Complex eye movement pattern biometrics: {A}nalyzing
  fixations and saccades}, in: \bibinfo{booktitle}{Proceedings of the
  International Conference on Biometrics}.
%Type = Inproceedings
\bibitem[{Holland and Komogortsev(2011)}]{Holland2011}
\bibinfo{author}{Holland, C.}, \bibinfo{author}{Komogortsev, O.V.},
  \bibinfo{year}{2011}.
\newblock \bibinfo{title}{Biometric identification via eye movement scanpaths
  in reading}, in: \bibinfo{booktitle}{2011 International Joint Conference on
  Biometrics {(IJCB)}}, pp. \bibinfo{pages}{1--8}.
%Type = Inproceedings
\bibitem[{Jaakkola and Haussler(1999)}]{jaakkola1999}
\bibinfo{author}{Jaakkola, T.}, \bibinfo{author}{Haussler, D.},
  \bibinfo{year}{1999}.
\newblock \bibinfo{title}{Exploiting generative models in discriminative
  classifiers}, in: \bibinfo{booktitle}{Advances in neural information
  processing systems}, pp. \bibinfo{pages}{487--493}.
%Type = Article
\bibitem[{Juhola et~al.(2013)Juhola, Zhang and Rasku}]{Juhola2013}
\bibinfo{author}{Juhola, M.}, \bibinfo{author}{Zhang, Y.},
  \bibinfo{author}{Rasku, J.}, \bibinfo{year}{2013}.
\newblock \bibinfo{title}{Biometric verification of a subject through eye
  movements}.
\newblock \bibinfo{journal}{Computers in Biology and Medicine}
  \bibinfo{volume}{43}, \bibinfo{pages}{42--50}.
%Type = Phdthesis
\bibitem[{Kasprowski(2004)}]{Kasprowski2004}
\bibinfo{author}{Kasprowski, P.}, \bibinfo{year}{2004}.
\newblock \bibinfo{title}{Human identification using eye movements}.
\newblock Ph.D. thesis. Silesian Unversity of Technology, Poland.
%Type = Inproceedings
\bibitem[{Kasprowski and Ober(2004)}]{KasprowskiOber2004}
\bibinfo{author}{Kasprowski, P.}, \bibinfo{author}{Ober, J.},
  \bibinfo{year}{2004}.
\newblock \bibinfo{title}{Eye movements in biometrics}, in:
  \bibinfo{booktitle}{{International Workshop on Biometric Authentication}},
  pp. \bibinfo{pages}{248--258}.
%Type = Inproceedings
\bibitem[{Kinnunen et~al.(2010)Kinnunen, Sedlak and Bednarik}]{Kinnunen2010}
\bibinfo{author}{Kinnunen, T.}, \bibinfo{author}{Sedlak, F.},
  \bibinfo{author}{Bednarik, R.}, \bibinfo{year}{2010}.
\newblock \bibinfo{title}{Towards task-independent person authentication using
  eye movement signals}, in: \bibinfo{booktitle}{Proceedings of the 2010
  Symposium on Eye-Tracking Research and Applications {(ETRA `10)}}, pp.
  \bibinfo{pages}{187--190}.
%Type = Inproceedings
\bibitem[{Kumar et~al.(2007)Kumar, Garfinkel, Boneh and Winograd}]{Kumar2007}
\bibinfo{author}{Kumar, M.}, \bibinfo{author}{Garfinkel, T.},
  \bibinfo{author}{Boneh, D.}, \bibinfo{author}{Winograd, T.},
  \bibinfo{year}{2007}.
\newblock \bibinfo{title}{Reducing shoulder-surfing by using gaze-based
  password entry}, in: \bibinfo{booktitle}{Proceedings of the 3rd Symposium on
  Usable Privacy and Security}, pp. \bibinfo{pages}{13--19}.
%Type = Inproceedings
\bibitem[{Landwehr et~al.(2014)Landwehr, Arzt, Scheffer and
  Kliegl}]{Landwehr2014}
\bibinfo{author}{Landwehr, N.}, \bibinfo{author}{Arzt, S.},
  \bibinfo{author}{Scheffer, T.}, \bibinfo{author}{Kliegl, R.},
  \bibinfo{year}{2014}.
\newblock \bibinfo{title}{A model of individual differences in gaze control
  during reading}, in: \bibinfo{booktitle}{EMNLP}, pp.
  \bibinfo{pages}{1810--1815}.
%Type = Inproceedings
\bibitem[{Maeder et~al.(2004)Maeder, Fookes and Sridharan}]{Maeder2004}
\bibinfo{author}{Maeder, A.}, \bibinfo{author}{Fookes, C.},
  \bibinfo{author}{Sridharan, S.}, \bibinfo{year}{2004}.
\newblock \bibinfo{title}{Gaze based user authentication for personal computer
  applications}, in: \bibinfo{booktitle}{Proceedings of the 2004 International
  Symposium on Intelligent Multimedia, Video and Speech Processing}, pp.
  \bibinfo{pages}{727--730}.
%Type = Inproceedings
\bibitem[{Makowski et~al.(2018)Makowski, J\"ager, Abdelwahab, Landwehr and
  Scheffer}]{Makowski2018}
\bibinfo{author}{Makowski, S.}, \bibinfo{author}{J\"ager, L.A.},
  \bibinfo{author}{Abdelwahab, A.}, \bibinfo{author}{Landwehr, N.},
  \bibinfo{author}{Scheffer, T.}, \bibinfo{year}{2018}.
\newblock \bibinfo{title}{A discriminative model for identifying readers and
  assessing text comprehension from eye movements}, in:
  \bibinfo{booktitle}{Proceedings of the European Conference on Machine
  Learning (ECML)}.
%Type = Article
\bibitem[{Noton and Stark(1971)}]{Noton1971}
\bibinfo{author}{Noton, D.}, \bibinfo{author}{Stark, L.}, \bibinfo{year}{1971}.
\newblock \bibinfo{title}{Scanpaths in eye movements during pattern
  perception}.
\newblock \bibinfo{journal}{Science} \bibinfo{volume}{171},
  \bibinfo{pages}{308--311}.
%Type = Article
\bibitem[{Rigas et~al.(2012)Rigas, Economou and Fotopoulos}]{Rigas2012}
\bibinfo{author}{Rigas, I.}, \bibinfo{author}{Economou, G.},
  \bibinfo{author}{Fotopoulos, S.}, \bibinfo{year}{2012}.
\newblock \bibinfo{title}{Biometric identification based on the eye movements
  and graph matching techniques}.
\newblock \bibinfo{journal}{Pattern Recogntion Letters} \bibinfo{volume}{33},
  \bibinfo{pages}{786--792}.
%Type = Article
\bibitem[{Rigas et~al.(2016)Rigas, Komogortsev and Shadmehr}]{Rigas2016}
\bibinfo{author}{Rigas, I.}, \bibinfo{author}{Komogortsev, O.},
  \bibinfo{author}{Shadmehr, R.}, \bibinfo{year}{2016}.
\newblock \bibinfo{title}{Biometric recognition via eye movements: Saccadic
  vigor and acceleration cues}.
\newblock \bibinfo{journal}{{ACM} Transactions on Applied Perception}
  \bibinfo{volume}{13}, \bibinfo{pages}{6}.
%Type = Article
\bibitem[{Sch{\"u}tt et~al.(2017)Sch{\"u}tt, Rothkegel, Trukenbrod, Reich,
  Wichmann and Engbert}]{schutt2017likelihood}
\bibinfo{author}{Sch{\"u}tt, H.H.}, \bibinfo{author}{Rothkegel, L.O.},
  \bibinfo{author}{Trukenbrod, H.A.}, \bibinfo{author}{Reich, S.},
  \bibinfo{author}{Wichmann, F.A.}, \bibinfo{author}{Engbert, R.},
  \bibinfo{year}{2017}.
\newblock \bibinfo{title}{Likelihood-based parameter estimation and comparison
  of dynamical cognitive models.}
\newblock \bibinfo{journal}{Psychological review} \bibinfo{volume}{124},
  \bibinfo{pages}{505}.
%Type = Inproceedings
\bibitem[{Silver and Biggs(2006)}]{SilverBiggs2006}
\bibinfo{author}{Silver, D.L.}, \bibinfo{author}{Biggs, A.},
  \bibinfo{year}{2006}.
\newblock \bibinfo{title}{Keystroke and eye-tracking biometrics for user
  identification}, in: \bibinfo{booktitle}{Proceedings of the 2006
  International Conference on Artificial Intelligence ({ICAI} 2006)}, pp.
  \bibinfo{pages}{344--348}.
%Type = Inproceedings
\bibitem[{Srivastava et~al.(2015)Srivastava, Agrawal, Roy and
  Tiwary}]{Srivastava2015}
\bibinfo{author}{Srivastava, N.}, \bibinfo{author}{Agrawal, U.},
  \bibinfo{author}{Roy, S.}, \bibinfo{author}{Tiwary, U.S.},
  \bibinfo{year}{2015}.
\newblock \bibinfo{title}{Human identification using linear multiclass svm and
  eye movement biometrics}, in: \bibinfo{booktitle}{8th International
  Conference on Contemporary Computing {(IC3)}}, pp. \bibinfo{pages}{365--369}.
%Type = Inproceedings
\bibitem[{Weaver et~al.(2011)Weaver, Mock and Hoanca}]{Weaver2011}
\bibinfo{author}{Weaver, J.}, \bibinfo{author}{Mock, K.},
  \bibinfo{author}{Hoanca, B.}, \bibinfo{year}{2011}.
\newblock \bibinfo{title}{Gaze-based password authentication through automatic
  clustering of gaze points}, in: \bibinfo{booktitle}{2011 {IEEE} International
  Conference on Systems, Man, and Cybernetics {(SMC)}}, pp.
  \bibinfo{pages}{2749--2754}.
%Type = Inproceedings
\bibitem[{Yoon et~al.(2014)Yoon, Carmichael and Tourassi}]{Yoon2014}
\bibinfo{author}{Yoon, H.J.}, \bibinfo{author}{Carmichael, T.R.},
  \bibinfo{author}{Tourassi, G.}, \bibinfo{year}{2014}.
\newblock \bibinfo{title}{Gaze as a biometric}, in:
  \bibinfo{booktitle}{Proceedings of the 2014 {SPIE} Medical Imaging
  Conference: {I}mage Perception, Observer Performance, and Technology
  Assessment}.
%Type = Article
\bibitem[{Zhang et~al.(2014)Zhang, Laurikkala and Juhola}]{Zhang2014}
\bibinfo{author}{Zhang, Y.}, \bibinfo{author}{Laurikkala, J.},
  \bibinfo{author}{Juhola, M.}, \bibinfo{year}{2014}.
\newblock \bibinfo{title}{Biometric verification of a subject with eye
  movements, with special reference to temporal variability in saccades between
  a subject's measurements}.
\newblock \bibinfo{journal}{International Journal of Biometrics}
  \bibinfo{volume}{6}, \bibinfo{pages}{75--94}.

\end{thebibliography}

%\clearpage

%%%% This page is for instructions only, once the article is finalize please omit the below text before creating the final PDF
%\normalMode

%\section*{Instructions to Authors for LaTeX template:}
%
%\section{ZIP mode for LaTeX template:}
%
%The zip package is created as per the guide lines present on the URL http://www.elsevier.com/author-schemas/ preparing-crc-journal-articles-with-latex for creating the LaTeX zip file of Procedia LaTeX template.  The zip generally contains the following files:
%\begin{Itemize}[]\leftskip-17.7pt\labelsep3.3pt
%\item ecrc.sty
%\item  elsarticle.cls
%\item elsdoc.pdf
%\item .bst file
%\item Manuscript templates for use with these bibliographic styles
%\item  Generic and journal specific logos, etc.
%\end{Itemize}
%
%The LaTeX package is the main LaTeX template. All LaTeX support files are required for LaTeX pdf generation from the LaTeX template package. 
%
%{\bf Reference style .bst file used for collaboration support:} In the LaTeX template packages of all Procedia titles a new ``.bst'' file is used which supports collaborations downloaded from the path http://www.elsevier.com/author-schemas/the-elsarticle-latex-document-class
%
%\section{Reference style used in Computer Science:}
%\let\footnotesize\normalsize
%\hspace*{-10pt}\begin{tabular*}{\hsize}{@{}ll@{}}
%{\bf Title}&{\bf Reference style} \\[6pt]
%PROCS  & 3 Vancouver Numbered
%\end{tabular*}

\end{document}